%% file: main.tex
\documentclass[11pt]{article}
\usepackage{etoolbox}

\newtoggle{arxiv}
\toggletrue{arxiv}
\input{acl-preamble}

\title{\oursshort: A Linguistically Diverse Dataset for 3D Visual Grounding}

\author{Austin T.~Wang$^1$ \quad ZeMing Gong$^1$ \quad Angel X. Chang$^{1,2}$\\
Simon Fraser University$^1$ \quad Alberta Machine Intelligence Institute (Amii)$^2$\\
{\tt\small \{atw7, zmgong, angelx\}@sfu.ca}\\
{\small{\url{https://3dlg-hcvc.github.io/vigil3d/}}}
\\
}

\begin{document}

\maketitle

\input{section/paper/00_abstract}
\input{section/paper/01_introduction}
\input{section/paper/02_related_work}
\input{section/paper/03_dataset_analysis}
\input{section/paper/04_dataset}

\input{section/paper/05_experiments}
\input{section/paper/06_conclusion}
\vfill\eject  
\input{section/paper/07_limitations}
\iftoggle{arxiv}{
  \input{section/paper/09_acknowledgements}
}{}

{
    \small
    \bibliography{main}
}

\vfill\eject  
\input{section/paper/08_supplement}

\end{document}

%% file: acl-preamble.tex
\pdfoutput=1

\iftoggle{arxiv}{
  \usepackage[preprint]{styles/acl/acl}
}{
  \usepackage[review]{styles/acl/acl}
}

\input{macros}

\usepackage{times}
\usepackage{latexsym}
\usepackage{booktabs}

\usepackage[T1]{fontenc}

\usepackage[utf8]{inputenc}

\usepackage{microtype}

\usepackage{inconsolata}

\usepackage{graphicx}

%% file: macros.tex
\usepackage{multirow}
\usepackage{colortbl}
\usepackage{wrapfig}
\usepackage{xspace}
\usepackage{amssymb, amsmath, mathtools}
\usepackage{tablefootnote}
\usepackage{listings}
\usepackage{cleveref}

\def\Snospace~{\S{}}
\newcommand{\appref}[1]{\hyperref[#1]{Appendix~\ref*{#1}}}

\DeclareCaptionType[placement={!ht}]{listing}[Listing][Code Listings]

\DeclarePairedDelimiter\abs{\lvert}{\rvert} 

\usepackage{pifont}
\newcommand{\xmark}{\ding{55}}
\definecolor{vlightgray}{gray}{0.87}
\newcommand{\myxmark}{{\color{vlightgray}\xmark{}}}
\newcommand{\mycheckmark}{\checkmark}
\newcommand\doublecheck{\checkmark\kern-0.6em\checkmark}
\newcommand\triplecheck{\checkmark\kern-0.6em\checkmark\kern-0.6em\checkmark}
\newcommand{\doublecheckmark}{$\triplecheck$}

\definecolor{LightCyan}{rgb}{0.88,1,1}
\newcolumntype{g}{>{\columncolor{LightCyan}}r}

\expandafter\def\expandafter\normalsize\expandafter{%
    \normalsize%
    \setlength\abovedisplayskip{2pt}%
    \setlength\belowdisplayskip{2pt}%
    \setlength\abovedisplayshortskip{-8pt}%
    \setlength\belowdisplayshortskip{2pt}%
}

\definecolor{tableaublue}{rgb}{0.31, 0.47, 0.65}
\definecolor{tableauorange}{rgb}{0.95, 0.56, 0.17}
\definecolor{tableaured}{rgb}{0.88, 0.34, 0.35}
\definecolor{tableaucyan}{rgb}{0.46, 0.72, 0.70}
\definecolor{tableaugreen}{rgb}{0.35, 0.63, 0.31}

\newcommand{\oursshort}{ViGiL3D\xspace}
\newcommand{\oursintro}{\textbf{Vi}sual \textbf{G}rounding with D\textbf{i}verse \textbf{L}anguage in \textbf{3D} (\oursshort)\xspace}

\newcommand{\best}[1]{\textbf{#1}}

\newcommand{\abbr}[1]{\textbf{#1}}

\newcommand{\TODO}[1]{\textcolor{red}{[TODO: #1]}}
\newcommand{\angel}[1]{\textcolor{orange}{\textbf{Angel:} #1}}
\newcommand{\austin}[1]{\textcolor{violet}{\textbf{Austin:} #1}}
\newcommand{\ming}[1]{\textcolor{purple}{\textbf{ZeMing:} #1}}
\newcommand{\fromreviews}[1]{\textcolor{olive}{#1}}

\newcommand{\cameraready}[1]{\textcolor{tableauorange}{#1}}

\renewcommand{\TODO}[1]{}
\renewcommand{\angel}[1]{}
\renewcommand{\ming}[1]{}
\renewcommand{\austin}[1]{}
\renewcommand{\fromreviews}[1]{#1}

\renewcommand{\cameraready}[1]{#1}

\newcommand{\vlmlimitations}[1]{\textcolor{tableaublue}{#1}}
\newcommand{\thresholdselect}[1]{\textcolor{tableauorange}{#1}}
\newcommand{\promptvalidation}[1]{\textcolor{tableaured}{#1}}
\newcommand{\statistical}[1]{\textcolor{tableaucyan}{#1}}
\newcommand{\reweightedeval}[1]{\textcolor{tableaugreen}{#1}}

\renewcommand{\vlmlimitations}[1]{#1}
\renewcommand{\thresholdselect}[1]{#1}
\renewcommand{\promptvalidation}[1]{#1}
\renewcommand{\statistical}[1]{#1}
\renewcommand{\reweightedeval}[1]{#1}

%% file: section/paper/00_abstract.tex
\begin{abstract}
    3D visual grounding (3DVG) involves localizing entities in a 3D scene referred to by natural language text. Such models are useful for embodied AI and scene retrieval applications, which involve searching for objects or patterns using natural language descriptions. 
    While recent works have focused on LLM-based scaling of 3DVG datasets, these datasets do not capture the full range of potential prompts which could be specified in the English language. 
    To ensure that we are scaling up and testing against a useful and representative set of prompts, we propose a framework for linguistically analyzing 3DVG prompts and introduce \oursintro, a diagnostic dataset for evaluating visual grounding methods against a diverse set of language patterns.
    We evaluate existing open-vocabulary 3DVG methods to demonstrate that these methods are not yet proficient in understanding and identifying the targets of more challenging, out-of-distribution prompts, toward real-world applications.
        

\end{abstract}

%% file: section/paper/01_introduction.tex
\section{Introduction}
\label{sec:intro}

\input{figs/teaser}

  Given a natural language description and a 3D scene, 3D visual grounding (3DVG) models localize the target entities in the scene described by the prompt. 
  The ability to locate objects in 3D scenes based on language is useful for a variety of applications in computer graphics, robotics, and dialogue with virtual and augmented reality assistants.
  \emph{Open-vocabulary} models, specifically, can generalize to novel object classes not seen during training. Such novel classes may appear in the text corpus used to pretrain the language model but are not part of the 3DVG training set. 
  Building high performance visual grounding models enables downstream applications in embodied AI, such as robots identifying objects in an environment, and large-scale 3D scene retrieval, such as searching interior design databases for objects or attributes.

  Compared to the success of recent 2D vision-language foundation models, 
  progress in 3DVG has been slow due to the lack of large-scale 3D datasets paired with language. The most commonly used datasets---ScanRefer~\cite{chen2020scanrefer}, Nr3D, and Sr3D~\cite{achlioptas2020referit3d}---are based on just $\sim$700 scenes and $\sim$170K prompts combined. 
  To alleviate the lack of data, recent work combined existing scene datasets and constructed LLM-based pipelines for scaling up grounding annotations~\cite{yang20243d, zhu20233d, jia2024sceneverse}. 
  While these methods achieve reasonable performance on ScanRefer and similar datasets, these benchmarks do not fully evaluate how well models handle the diversity of language patterns and types of grounding prompts found in the English language. It is of value to ensure that we are scaling a complete set of grounding prompts and evaluating on a dataset that accurately measures understanding of language and vision and demonstrates their viability in real-world situations.

  We thus propose \oursintro, a diagnostic 3DVG dataset for evaluating visual grounding methods against a diverse range of language patterns, in order to determine 1) how well existing methods actually perform and 2) their specific strengths and weaknesses in grounding targets based on different linguistic phenomena. 
  \cameraready{We develop an automated method to analyze 3DVG datasets, identifying} a lack of linguistic diversity and several infrequently represented language patterns, such as negations. \cameraready{While there are methods to extract attributes and relationships from scenes and descriptions \citep{sun20243d, qian2024nuscenes, gu2024conceptgraphs}, ours also categorizes them and further identifies high-level linguistic patterns not captured in scene graphs.}
  Given the limitations of existing datasets, we manually annotate \oursshort as a test dataset \cameraready{with an emphasis on prompt diversity} and benchmark prior 3DVG models, demonstrating that the best models achieve an accuracy at least 20 points lower than their respective ScanRefer performances. 
  We analyze the performance of each method on subgroups of language patterns to draw important insights about where the models are succeeding or falling short, demonstrating that further work is needed to bridge the gap of translating language understanding to 3D. 
  Simply scaling data volume alone is insufficient for achieving good performance across the entire domain of 3DVG, but rather the right distribution must be captured.
  We believe that our work will contribute toward understanding the true performance and limits of the state-of-the-art and move us toward scaling the right types of prompts and building models which can tackle problems in real-world applications.
  \cameraready{In summary:
  \begin{itemize}
      \item we propose an automated pipeline for analyzing linguistic patterns in visual grounding descriptions and use it to investigate the limitations of existing 3DVG datasets;
      \item we construct a new dataset, \oursshort, for evaluating 3DVG methods against more challenging and diverse grounding descriptions than existing datasets; and
      \item we show that current 3DVG models perform worse on \oursshort than existing benchmarks, demonstrating the value of \oursshort for future 3DVG model development.
  \end{itemize}
  }


%% file: figs/teaser.tex
  \begin{figure}[ht]
  \centering
  \includegraphics[trim={0 0 0 5px},clip,width=\linewidth]{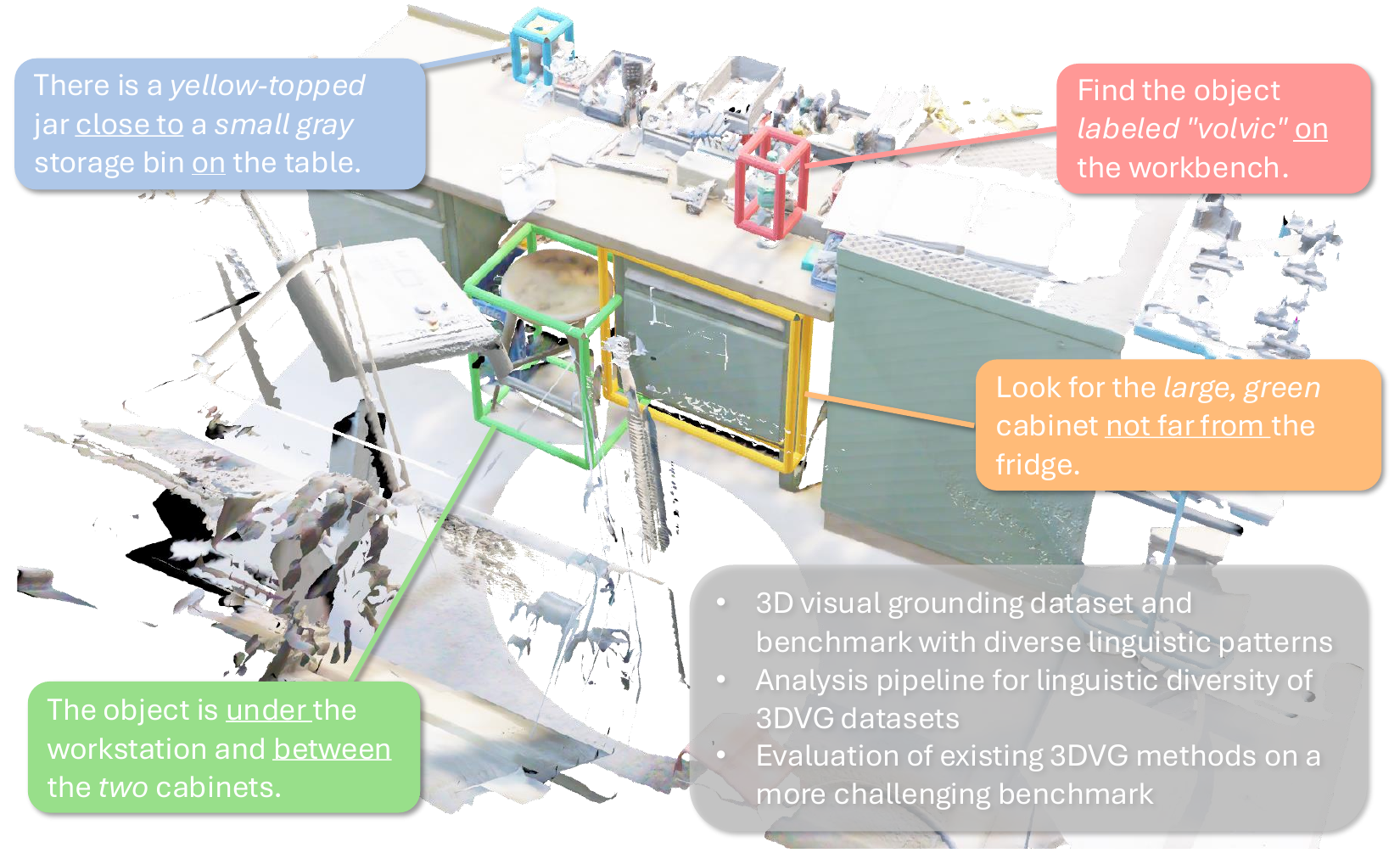}
\captionof{figure}{
  \textit{Overview of \oursshort.} We propose a new dataset for visual grounding to better evaluate 3D visual grounding methods on the wide diversity of linguistic patterns possible to refer to objects in a scene. While existing datasets largely contain more homogenous and direct prompts, \oursshort includes coarse-grained object references, negation, reference resolution, and other phenomena in more varied sentence structures which allow us to more precisely and comprehensively measure state-of-the-art performance.
}
  \label{fig:teaser}
 \vspace{-4mm}
\end{figure}

%% file: section/paper/02_related_work.tex
\section{Related Work}

\textbf{3DVG Datasets.} Existing datasets differ by scene type (indoor vs. outdoor and types of rooms), acquisition of 3D data (real-world vs. synthetic) and text (annotation process), scene size (single room vs. multiple rooms), and the scale and diversity of object or language annnotations. 
3DVG research has primarily focused on providing language prompts for indoor scene datasets,  with limited datasets for outdoor 3DVG~\cite{miyanishi2023cityrefer}.
  
Early datasets obtained language prompts from crowdworkers~\cite{chen2020scanrefer, achlioptas2020referit3d}, or using simple templates~\cite{achlioptas2020referit3d} on ScanNet~\cite{dai2017scannet}, a dataset of real-world indoor rooms with semantically annotated objects.
  Later VG datasets used other sources of real-world 3D scene data~\cite{kato2023arkitscenerefer}, as well as synthetic datasets. 
  Recent work has applied captioning models, LLMs, and scene graph generation methods to automatically generate prompts on aggregate scene datasets~\cite{jia2024sceneverse, yang20243d, zhu20233d, hong20233d, huang2023embodied, li2023m3dbench, lyu2024mmscan, yang20243d, zhang2024vla}. 
  
Other efforts provided denser alignment~\cite{abdelreheem2024scanents3d}, grounding without object names~\cite{wu2023eda}, and explored evaluation for grounding to multiple targets~\cite{zhang2023multi3drefer}, identifying regions or objects by function \cite{delitzas2024scenefun3d, he2024segpoint, zhang2024task}, or requiring reasoning to ground objects \cite{szymanska2024space3d, zhu2024empowering}. 
However, both manual and LLM-scaled 3DVG datasets only capture part of the diverse language patterns in real-world applications, resulting in limitations in performance when methods are tested on out-of-distribution prompts. We propose a new diagnostic dataset covering different linguistic phenomena to study the performance of 3DVG models.

\textbf{3DVG Methods.} Traditional models fuse independently extracted visual and text features to identify the most likely points or regions corresponding to a target \cite{chen2020scanrefer, achlioptas2020referit3d, abdelreheem2024scanents3d, wu2023eda, jain2022bottom, cai20223djcg, chen2023unit3d, jin2023context, chen2022language}. We focus on evaluating open-vocabulary methods, which generalize to a broader set of prompts and object classes than those used during the training or fine-tuning of grounding capabilities \cite{peng2023openscene, takmaz2023openmask3d, kerr2023lerf, yang2024llm, yuan2024visual}, enabled first by models such as CLIP \cite{radford2021learning} and later by large language models (LLMs) \cite{achiam2023gpt}. Recent work has focused on developing 3D foundation models for a wide variety of tasks on 3D scene data beyond visual grounding, including generic visual question-answering, captioning, segmentation, and similar tasks \cite{jia2024sceneverse, yang20243d, zhu20233d, hong20233d, huang2023embodied, li2023m3dbench, lyu2024mmscan, he2024segpoint, man2024lexicon, zhu2024unifying}. These methods are pretrained on LLM-scaled datasets to aid generalization. We show that these datasets lack crucial aspects of language, resulting in subpar performance of current foundation models on out-of-distribution 3DVG prompts.


%% file: section/paper/03_dataset_analysis.tex
  \input{table/analysis_criteria}

  \input{table/attributes}
  \input{table/spatial_relationships}

\section{Analysis of Prior Datasets}

We annotate prompts from prior visual grounding datasets to identify strengths and shortcomings of each with respect to their linguistic properties, and to better  understand the impact of the datasets on the methods trained and evaluated on them.

\subsection{Language Patterns}

We break down a grounding description into the target, anchors, attributes, and relationships. Each object is either a target (i.e. primary object of interest), or an anchor (i.e. an object or other reference region or agent used to help identify the location of the target). Attributes describe a target or anchor independent of the context, and relationships are used to compare two entities in the scene.
To characterize these four aspects, we devise a set of \cameraready{35 count-based or binary} metrics for analyzing 1) \textbf{language diversity} (DIV), or coverage of a variety of different types and patterns; 2) \textbf{language resolution} (RES), the ability to link descriptors with their referents; and 3) \textbf{understanding attributes and relationships} (UAR), the ability to correspond each constraint to their appearance in the scene.
\cameraready{In particular, these metrics track the reference types for targets and anchors, types and quantities of attributes and relationships, and language patterns such as negations. We also measure overall language diversity through token bigram frequency, similar to \cite{mensink2023encyclopedic}.}
Each of the criteria is documented in \autoref{tab:dataset-criteria}.
  
\input{figs/dataset_analysis}

To analyze each dataset, we devise an automated pipeline that assesses the occurrence of different language properties in each prompt (see \autoref{fig:dataset-analysis}).  
For each prompt, we use GPT-4o~\cite{achiam2023gpt} to extract an augmented scene graph that captures the objects, attributes, and relationships in the description.
We include the full prompts used for scene graph extraction in \autoref{appendix:dataset-analysis}. 
We also use SpaCy~\cite{spacy2} to obtain a dependency parse of the tokens to measure diversity of bigrams in the dataset.

To validate our pipeline, we compare its output against 225 manually annotated prompts randomly sampled from all of the datasets, including at least 20 prompts from each dataset.
Our pipeline achieves an average precision and recall across 28 measured binary metrics of 0.86 and 0.91, respectively.
The median error and median absolute deviation for each of the counts of attributes and relationships pertaining to the target and anchor objects was 0.0 in all cases, and the mean absolute error was around 0.43.
This shows the robustness of our pipeline for prompt analysis.
Details for the manual validation are in \autoref{appendix:dataset-analysis}.
  
\subsection{Datasets}
  
We select commonly used 3DVG datasets, as well as several LLM-scaled datasets given their importance in developing 3DVG foundation models. We evaluate \textbf{ScanRefer} \cite{chen2020scanrefer}, \textbf{Nr3D/Sr3D+} \cite{achlioptas2020referit3d}, and \textbf{Multi3DRefer} \cite{zhang2023multi3drefer}. Building on the crowdsourced ScanRefer, Nr3D and Sr3D+ used manual and template-based methods, respectively, to generate prompts focused on discriminating objects of the same class. Multi3DRefer introduced zero- and multi-target grounding objectives. \textbf{Instruct3D} \cite{he2024segpoint}, another human-annotated dataset, extends the ideas of Multi3DRefer with an emphasis on reasoning about the function of objects.
We further examine \textbf{ScanScribe} \cite{zhu20233d}, \textbf{3D-GRAND} \cite{yang20243d}, and \textbf{SceneVerse} \cite{jia2024sceneverse} as recent large-scale 3D datasets, largely leveraging template-based generation and GPT QA or rephrasing to generate prompts. Additional details for these datasets are in \autoref{appendix:dataset-analysis}.

\subsection{Analysis}

\input{table/dataset_comparison}
  
To compare datasets, we apply our automated pipeline to 1000 randomly sampled prompts from prior datasets and all 350 prompts from \oursshort. \cameraready{We show the most differentiating metrics in \autoref{tab:dataset-comparison}} 
and all others in \appref{appendix:additional-dataset-analysis}.
We identify several shortcomings across many existing datasets.

\textbf{Balanced number of descriptors.} 3D-GRAND and ScanScribe employ an excessive number of attributes or relationships to describe the target object (e.g., ``The chair is behind the desk, on the left side of the circular black table, to the right of the rectangular shelf, and to the left of the other chair.''). While this ensures the target is uniquely defined, it also provides the models with more information than they would be given in most practical grounding prompts. 3D-GRAND and ScanScribe, for instance, have more than three times the number of attributes per prompt compared to the manually annotated datasets. Furthermore, with many attribute and relationship types, the model can over-rely on certain signals and ignore others. Sr3D and SceneVerse, on the other hand, have very few descriptors and thus may not have sufficient diversity to represent more complex prompts. 

\textbf{Overly specific target references.} In most datasets, the target object is referenced by its explicit class. This makes it easier to identify the target when it is unique in the scene, allowing models primarily to attend to the object class name and ignore other signals. While \citet{wu2023eda} evaluates this scenario, they only mask with the ``object'' keyword, whereas there are further gradations of detail that can be represented. Instruct3D \cite{he2024segpoint} requires the model to identify objects based on described function and reasoning rather than semantic class. However, its scope is relatively narrow with respect to function, ignoring other potential attribute or relationship types.

\textbf{Missing negations}. Most prompts focus on the positive descriptors of the target object. However, properly eliminating objects based on any attributes that are \textit{not} true of the target is also important, as a significant aspect of language. While ScanScribe does include some negation, it is often not useful toward identifying the target (e.g., ``The description doesn't provide enough context to determine the location of the box with certainty.'', commonly found in similar forms in many prompts).

\textbf{Language diversity.} Most of the datasets have low proportions of unique lexical bigrams, with \oursshort significantly outpacing other datasets at 0.52. We further observe that the target is the first noun phrase in more than 80\% of prompts from all datasets except Instruct3D and \oursshort, further showing a lack of diversity in sentence structure in existing datasets.


\textbf{Underrepresented attribute or relationship types}. Some of the more challenging types are underrepresented, including numerical cues, object states, and complex spatial relationships involving long distances or multiple objects.
Text labels are particularly difficult because of insufficient point cloud resolution to capture writing.

Given these shortcomings, our goal is to develop a diverse diagnostic benchmark which can adequately represent these different language phenomena for evaluation. While combining existing datasets is a viable strategy, as we saw some linguistic phenomena are absent across all datasets. Furthermore, the representations of most phenomena, while technically present, do not adequately allow us to understand how well models parse them for 3D understanding, due to strong correlations between phenomena which can confound analysis.

%% file: table/analysis_criteria.tex
\begin{table*}[tb]
\centering
\resizebox{\linewidth}{!}
{
\begin{tabular}{@{}llp{4.2in}l@{}}
\toprule
Cat & Metric & Definition & Examples \\
\midrule
\multicolumn{4}{l}{\textbf{Attribute Understanding} (average number of attributes per prompts)} \\
\midrule
UAR & Total attributes (Attr-All) & entire prompt & A \textit{round} table sits in front of the \textit{white} sofa. \\
RES & Target attributes (Attr-Tgt) & for describing the target & These are the \textit{fancy, wooden} chairs in the room. \\
RES & Anchor attributes (Attr-Anc) & for describing the anchors & Where is the couch that is farthest from the \textit{largest} bookshelf? \\
DIV & Attribute Type & proportion of prompts with a specific attribute type describing an object & See \autoref{tab:attributes} \\
\midrule
\multicolumn{4}{l}{\textbf{Relationship Understanding} (average number of relationship per prompts)} \\
\midrule
UAR & Total relationships (Rel-All) & entire prompt & Look for a personal computer \textit{near} a simple office chair \textit{with} arms. \\
RES & Target relationships (Rel-Tgt) & relationships which compare a target to an anchor & The box is \textit{under} the table. \\
RES & Anchor relationships (Rel-Anc) & relationships which compare anchors to other anchors & Use the outlet under the window which is \textit{second-to-the-right} in the room. \\
DIV & Relationship Type & proportion of prompts with a specific relationship type comparing objects or other entities in the scene & See \autoref{tab:spatial-relationships} \\
\midrule
\multicolumn{4}{l}{\textbf{Target Reference} (proportion of prompts)} \\
\midrule
UAR & Generic References (Gen) & target is referred to by a generic name (e.g. ``object'', ``thing'') & The \textit{object} is under the workstation and between the two cabinets. \\
UAR & Coarse-Grained References (CG) & target is referred to by a coarse category (e.g. ``appliance'', ``device'') & Locate the tallest \textit{appliance} in the kitchen. \\
UAR & Fine-Grained References (FG) & target is referred to by its specific category & Identify the stainless steel \textit{sink}. \\
RES & Coreferences (Cor) & coreferences is used to refer to objects & This is the rectangular whiteboard. \textit{It} is left of the other one. \\
RES & Not First Noun Phrase (NFN) & target object is not the first noun phrase in the description & Facing the standalone whiteboard, grab the closest \textit{chair} right behind you. \\
\midrule
\multicolumn{4}{l}{\textbf{Anchor Type} (proportion of prompts describing different types of anchors)} \\
\midrule
UAR & Single-Object Anchors (Sing) & anchor references a single object & On the brown, wooden \textit{table} is a small, rectangular projector. \\
UAR & Multi-Object Anchors (Mul) & anchor references multiple objects & The backpack is in between \textit{two other backpacks}. \\
UAR & Non-Object Anchors (Non) & anchor references a room or region & In between the counter and table is a black trash can in the \textit{corner}. \\
UAR & Agent-Based Anchors (Agt) & ences an agent or viewpoint & If \textit{you} are sitting on the couch, this is the bag further to the right. \\
\midrule
\multicolumn{4}{l}{\textbf{Language Patterns}} \\
\midrule
UAR & Negation (Neg) & proportion of prompts with negation & Find the food storage which does \textit{not} have a green, rectangular object on top. \\
DIV & \fromreviews{Lexical} bigrams (2lex) & proportion of unique bigrams of \fromreviews{lexical} tokens in descriptions & 
\\
\bottomrule
\end{tabular}
}
\vspace{-2mm}
\caption{\textbf{Dataset Criteria}. Summary of the metrics computed across each dataset, including the metric category (\textit{Cat}), how the metric is computed, and valid example prompts from \oursshort. }
\label{tab:dataset-criteria}
\end{table*}

%% file: table/attributes.tex
\begin{table}[tb]
\centering
\resizebox{\linewidth}{!}
{
\begin{tabular}{@{}lp{3.0in}@{}}
\toprule
Type & Examples \\
\midrule
\abbr{Col}or  & \small{On top of the shelf is a \textit{red} and \textit{yellow} object.} \\
\abbr{Siz}e & \small{This is the \textit{tallest} wooden furniture in the room.} \\
\abbr{Sha}pe & \small{Find the large, \textit{rectangular} object with magnets on the front.} \\
\abbr{Num}ber & \small{This is the larger of the \textit{two} toolboxes near the piano.} \\
\abbr{Mat}erial  & \small{When facing the radiator, the \textit{metal} rail is directly on your left.} \\
\abbr{Fun}ction & \small{Opposite the room from the left whiteboard, there is a device \textit{for heating the room}.} \\
\abbr{Tex}ture & \small{These are all the long, \textit{soft} places in the room.} \\
\abbr{Sty}le & \small{This is a \textit{classy} queen-size bed, farthest from the door of the hotel room.} \\
Text \abbr{Lab}el & \small{Find the object \textit{labeled ``caution''}.} \\
\abbr{Sta}te & \small{Find the \textit{folded} chair closest to the door.} \\
\bottomrule
\end{tabular}
}
\caption{Attribute types analyzed in each prompt.}
\label{tab:attributes}
\end{table}

%% file: table/spatial_relationships.tex
\begin{table}[tb]
\centering
\resizebox{\linewidth}{!}
{
\begin{tabular}{@{}lp{3in}@{}}
\toprule
Type & Examples \\
\midrule
\abbr{Near} & \small{Next to the table \textit{closest} to the entrance, find all of the unfolded chairs.} \\
\abbr{Far} & \small{The table is the largest one \textit{far} from the door.} \\
\abbr{Dir}ectional & \small{A black bag is \textit{in front of} a white trash can on the floor.} \\
\abbr{Ver}tical & \small{\textit{Under} the counter are two plastic bins. Find me the taller one.} \\
\abbr{Cont}ain & \small{\textit{In} the center of the room is a hexagonal conference table for meetings.} \\
\abbr{Arr}angement  & \small{In the \textit{stack} of three boxes, this is the third one from the bottom.} \\
\abbr{Ord}inal & \small{In the row of chairs and tables against the wall, find the \textit{third chair from the left}.} \\
\abbr{Comp}arison & \small{This fancy rotating display is the one \textit{nearest} to the orange carpet.} \\
\bottomrule
\end{tabular}
}
\caption{Relationship types analyzed in each prompt.
}
\label{tab:spatial-relationships}
\end{table}

%% file: figs/dataset_analysis.tex
  \begin{figure*}[t]
  \centering
  \includegraphics[trim={0 0 0 4px},clip,width=\linewidth]{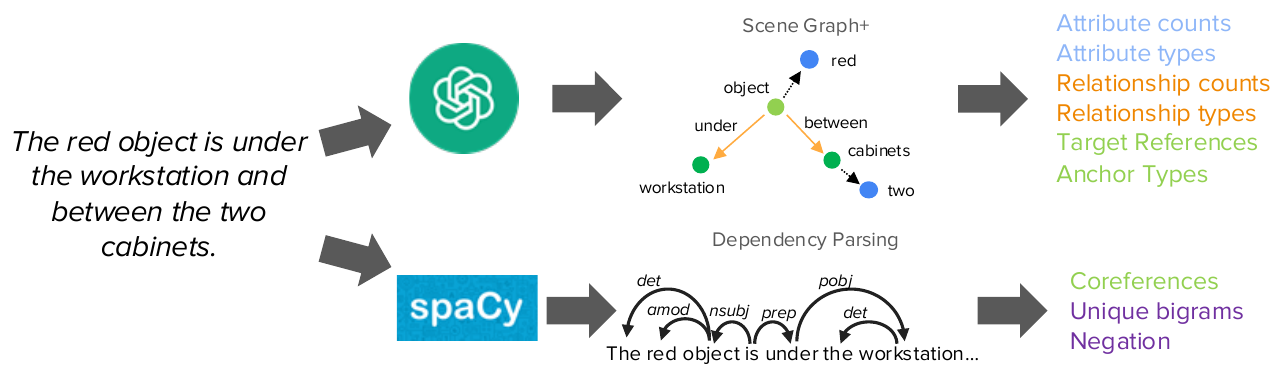}
  \caption{\textbf{Dataset Analysis Pipeline.} Using GPT-4o and SpaCy,  we automatically parse each visual grounding prompt and compute aggregate statistics for a variety of linguistic patterns for each dataset. We use GPT-4o for parsing an augmented scene graph optimized for visual grounding descriptions, and SpaCy for dependency parsing.
  }
  \label{fig:dataset-analysis}
\end{figure*}

%% file: table/dataset_comparison.tex
\begin{table*}[tb]
\centering
\resizebox{\linewidth}{!}
{
\begin{tabular}{@{}l cccccc ccccccc ccc ccc cc @{}}
\toprule
& \multicolumn{6}{c}{Attributes} & \multicolumn{7}{c}{Relationships} & \multicolumn{3}{c}{Target Reference} & \multicolumn{3}{c}{Anchor Type} & \multicolumn{2}{c}{Language} \\
\cmidrule(){2-7} \cmidrule(l){8-14} \cmidrule(l){15-17} \cmidrule(l){18-20} \cmidrule(l){21-22}
& All & Tgt & Anc & Num & Lab & State & All & Tgt & Anc & Far & Arr & Ord & Comp & Gen & CG & NFN & Mul & Non & Agt & Neg & 2lex \\
\midrule
ScanRefer & 1.90 & 1.21 & 0.68 & \mycheckmark & \myxmark & \myxmark & 2.33 & 1.89 & 0.44 & \mycheckmark & \mycheckmark & \mycheckmark & \myxmark & \myxmark & \myxmark & \myxmark & \mycheckmark & \doublecheckmark & \doublecheckmark & \myxmark & 0.20 \\
Nr3D & 1.16 & 0.64 & 0.52 & \doublecheckmark & \myxmark & \myxmark & 2.22 & 1.63 & 0.59 & \mycheckmark & \mycheckmark & \mycheckmark & \doublecheckmark & \myxmark & \myxmark & \doublecheckmark & \doublecheckmark & \doublecheckmark & \doublecheckmark & \mycheckmark & 0.27 \\
Sr3D+ & 0.05 & 0.02 & 0.03 & \myxmark & \myxmark & \myxmark & 1.00 & 1.00 & 0.00 & \doublecheckmark & \myxmark & \myxmark & \doublecheckmark & \myxmark & \myxmark & \myxmark & \myxmark & \myxmark & \myxmark & \myxmark & 0.02 \\
Multi3DRefer & 1.73 & 1.21 & 0.52 & \mycheckmark & \myxmark & \mycheckmark & 2.02 & 1.63 & 0.39 & \mycheckmark & \mycheckmark & \mycheckmark & \myxmark & \myxmark & \myxmark & \doublecheckmark & \mycheckmark & \doublecheckmark & \doublecheckmark & \myxmark & 0.28 \\
3D-GRAND & 5.81 & 4.68 & 1.12 & \mycheckmark & \myxmark & \myxmark & 2.81 & 2.71 & 0.10 & \mycheckmark & \mycheckmark & \myxmark & \mycheckmark & \myxmark & \myxmark & \myxmark & \doublecheckmark & \doublecheckmark & \doublecheckmark & \myxmark & 0.05 \\ 
ScanScribe & 6.04 & 1.15 & 4.89 & \myxmark & \myxmark & \myxmark & 3.55 & 3.35 & 0.21 & \myxmark & \myxmark & \myxmark & \doublecheckmark & \myxmark & \myxmark & \myxmark & \doublecheckmark & \mycheckmark & \doublecheckmark & \myxmark & 0.10 \\
SceneVerse & 0.41 & 0.35 & 0.07 & \myxmark & \myxmark & \myxmark & 1.33 & 1.30 & 0.03 & \mycheckmark & \myxmark & \myxmark & \myxmark & \myxmark & \myxmark & \mycheckmark & \mycheckmark & \myxmark & \myxmark & \myxmark & 0.16 \\ 
Instruct3D & 1.40 & 1.30 & 0.09 & \myxmark & \myxmark & \myxmark & 1.43 & 1.31 & 0.12 & \myxmark & \myxmark & \myxmark & \myxmark & \doublecheckmark & \doublecheckmark & \doublecheckmark & \myxmark & \doublecheckmark & \myxmark & \myxmark & 0.27 \\
 \rowcolor{gray!20} ViGiL3D & 1.62 & 1.09 & 0.53 & \doublecheckmark & \mycheckmark & \mycheckmark & 1.82 & 1.46 & 0.35 & \mycheckmark & \mycheckmark & \mycheckmark & \doublecheckmark & \mycheckmark & \mycheckmark & \doublecheckmark & \doublecheckmark & \doublecheckmark & \doublecheckmark & \mycheckmark & 0.45 \\ 
\bottomrule
\end{tabular}
}
\caption{\textbf{Dataset Comparison}. We show here a comparison of the linguistic differences of prompts from prior visual grounding datasets. Counts of attributes and relationships are shown as average counts, while binary metrics are reported as not present (\myxmark), some (\mycheckmark), or a lot (\doublecheckmark), as thresholded at 5\% and 20\% of the dataset sample, respectively. While most datasets skew toward having particular types of prompts, our \oursshort dataset (in gray) has the best coverage and diversity for all types of prompts and includes some prompt types not well-represented in most other datasets.
}
\label{tab:dataset-comparison}
\end{table*}


%% file: section/paper/04_dataset.tex
\section{\oursshort}

\input{table/our_dataset_overview}

We present \oursshort, a new diagnostic 3DVG dataset that captures a diversity of language patterns to assess how well recent 3DVG methods perform and where they fall short.

\input{table/result_scannet}
  
  We build our dataset on scenes from ScanNet \cite{dai2017scannet} and ScanNet++ \cite{yeshwanth2023scannet++}.
  We use ScanNet to assess the performance of prior works while controlling for the scene representation distribution and quality.
  We also annotate ScanNet++ to determine how well the model performance generalizes to new scenes and to leverage the higher quality 3D scenes, which may be critical for identifying smaller or more detailed targets. 

  To generate the prompts, annotators were asked to write grounding prompts for sampled objects in each scene given the RGB video stream and 3D point cloud. Targets were sampled from the ground truth annotations and could consist of zero, one, or multiple objects. Each prompt included a variety of language criteria and diverse but natural phrasing. Prompts were also designed to target a balanced level of specificity, in order to avoid both ambiguity and extraneous constraints on the target object. \cameraready{For zero-target prompts, annotators were instructed to craft them similarly to single-target prompts of objects in the scene but with modifications to make them unapplicable. This makes them more realistic and challenging for models compared to descriptions of absent object classes, as in Multi3DRefer~\citep{zhang2023multi3drefer}.}

  In total, we generate 350 prompts over 26 scenes for evaluation. Despite the small number, we are still able to demonstrate useful trends in the performance and believe that the principles of our dataset can be scaled up for future training and evaluation. More detailed statistics for \oursshort are provided in \autoref{appendix:dataset}.

  \cameraready{We compare the linguistic diversity of \oursshort to previous datasets in \autoref{tab:dataset-comparison}.
  Although many of the language patterns are captured by one of the prior datasets, none of them cover all of the patterns. For rare properties, such as text labels or negation, the datasets which do include them are fairly specialized or limited in other properties. Instruct3D, for example, is the only dataset with generic or coarse-grained target references but focuses primarily on grounding object functionality, at the exclusion of other attributes. 
  Combining or subsampling all existing datasets for evaluation would still be suboptimal compared to \oursshort for evaluation. 
  Combining datasets may induce correlations between patterns, such as generic target references always occurring in Instruct3D prompts with functionality-based attributes. Patterns in \oursshort, on the other hand, are annotated from the same distribution of prompts.}
  
  \cameraready{Additionally, \oursshort has more challenging prompts that require understanding of each phrase in the description and careful matching against objects in the scene. For instance, many of the descriptions in existing datasets are over-constrained, allowing models to ignore certain constraints (e.g. a model might only need to focus on the color of ``a white chair to the left of the fridge''). Zero-target prompts are similar to valid descriptions of objects in the scene, rather than describing object classes not present in the scene.
  Evaluation of \oursshort under a reweighted pattern distribution similar to ScanRefer's can be found in \appref{appendix:results} to demonstrate the difficulty and value of \oursshort beyond its diversity.
  }

%% file: table/our_dataset_overview.tex
\begin{table}[tb]
\centering
\resizebox{\linewidth}{!}
{
\begin{tabular}{@{}ll@{}}
\toprule
Metric & \oursshort \\
\midrule
Scene datasets & ScanNet, ScanNet++ \\
\# of prompts & 350 \\
\# of scenes & 35 \\
\midrule
Vocab size & 942 \\
\# sentences per prompt & 1.2 \\
Average prompt length & 14.1 \\
\midrule
\# prompts with 0 targets & 43 \\
\# prompts with 1 target & 275 \\
\# prompts with multiple targets & 32 \\
\bottomrule
\end{tabular}
}
\caption{\textbf{Statistics of \oursshort.}}
\label{tab:dataset-overview}
\end{table}

%% file: table/result_scannet.tex
\begin{table*}[tb]
\centering
\resizebox{\linewidth}{!}
{
\begin{tabular}{@{}l rrr rrr rr@{}}
\toprule
& \multicolumn{6}{c}{ViGiL3D} & \multicolumn{2}{c}{ScanRefer} \\
\cmidrule(){2-7} \cmidrule(l){8-9} 
 & Acc/GT & Acc@25 & Acc@50 & F1/GT & F1@25 & F1@50 & Acc@25 & Acc@50 \\
\midrule
OpenScene & 2.1 & 1.7 & 1.3 & 2.1 & 1.7 & 1.2 & 13.2 & 6.5 \\
LERF & 2.5 & 2.1 & 2.1 & 2.5 & 2.1 & 2.1 & 4.8 & 0.9 \\
ZSVG3D & 18.9 & 8.5 & 5.6 & 12.2 & 6.7 & 5.8 & 36.4\textsuperscript{*} & 32.7\textsuperscript{*} \\
LLM-Grounder & 2.5 & 7.1 & 5.0 & 2.5 & 5.3 & 3.1 & 17.1 & 5.3 \\
\midrule
3D-VisTA & 14.2 & \best{15.8} & \best{13.3} & 14.1 & \best{15.7} & \best{13.2} & 50.6 & 45.8 \\
3D-GRAND & 17.9 & \best{15.8} & 12.5 & 17.9 & 15.3 & 11.8 & 38.0 & 27.4 \\
PQ3D & \best{26.2} & 10.8 & 10.8 & \best{26.8} & 5.6 & 5.1 & \best{57.0} & \best{51.2} \\
\bottomrule
\end{tabular}
}
\caption{Accuracy and F1 score (\%) on \oursshort for ScanNet scenes. Each metric is computed using GT boxes or predicted boxes using IoU thresholds of 0.25 and 0.50, as is typical in the 3DVG literature. We compare against the overall ScanRefer validation set as a baseline, as reported by each method, to demonstrate the significant drop in performance on our prompts compared to existing datasets. 
\textsuperscript{*}ZSVG3D ScanRefer results use GPT-3.5, as opposed to GPT-4o on \oursshort.}
\label{tab:results-scannet}
\end{table*}


%% file: section/paper/05_experiments.tex
\input{table/result_subgroup}
\input{table/result_scannetpp}

\section{Experiments}
  
We apply recent 3DVG models on \oursshort and analyze their performance.

\subsection{3DVG models}
  
We focus on open-vocabulary methods that are designed to scale to new scene datasets and language descriptions not present in the 3DVG training data.
We consider three groups of methods: those that use CLIP to obtain a language-aware 3D representation, zero-shot 3DVG with LLMs, and methods trained on 3DVG data. 
\textit{CLIP aligned 3D representations}: We select \textbf{OpenScene} \cite{peng2023openscene} that projects features directly to point clouds, and \textbf{LERF} \cite{kerr2023lerf} that uses neural radiance fields. 
  \textit{Zero-shot with LLMs}: \textbf{ZSVG3D} \cite{yuan2024visual} and \textbf{LLM-Grounder} \cite{yang2024llm} both use LLMs for reasoning combined with independent localization modules, the former through program synthesis and the latter directly in natural language. 
  \textit{Trained with 3DVG data}: 
  \textbf{3D-VisTA} \cite{zhu20233d} and \textbf{3D-GRAND} \cite{yang20243d} are both transformer architectures trained on LLM-scaled datasets, thus allowing us to study the impacts of large-scale datasets on downstream performance and generalization. Lastly, \textbf{PQ3D} \cite{zhu2024unifying} is a representative promptable query-based model trained on an aggregate of many existing 3DVG datasets. Details of the configurations of each method are in \appref{appendix:baselines}.

  We ran each method with both ground truth boxes and boxes predicted from Mask3D \cite{schult2023mask3d}, following prior work. This enabled us to control for different methods of clustering points into objects and to analyze both the best-case performance as well as the realistic inference-level performance. Evaluating all methods required 22 GPU-hours on an RTX 4090 GPU.

  To evaluate grounding performance, we report accuracy and F1 for ground truth and predicted boxes using Mask3D. For Mask3D predictions, we use IoU thresholds of 0.25 and 0.50 following prior work~\cite{chen2020scanrefer, achlioptas2020referit3d, zhang2023multi3drefer}. For multi-target descriptions, we define accuracy as localizing all of the boxes within the specified IoU and the F1 score as the harmonic mean of precision and recall across objects.

  \input{figs/examples}

  \subsection{Results}
  
  Consistently across different methods, we observe that performance on \oursshort is significantly lower than benchmarked grounding results for ScanRefer, even for the same scenes. \autoref{tab:results-scannet} shows that even with the best model, PQ3D, the F1 is 24.4 points lower on our dataset, with similar trends for the other methods. This suggests that our prompts are likely out of distribution and harder than prior datasets. LLM-based methods and those trained on large datasets achieve significantly better performance compared to CLIP-based methods, likely due to the inability of CLIP to parse complex language patterns \cite{yuksekgonul2023and}. Howevercorrelation with ScanRefer performance is loose, with ZSVG3D and 3D-GRAND both outperforming 3D-VisTA with GT boxes.

  \textbf{GT vs. predicted boxes}. While better performance with ground truth information is expected, LLM-Grounder and 3D-VisTA actually score better with Mask3D predictions, and the performance of PQ3D drops precipitously compared to 3D-VisTA and 3D-GRAND, both of which achieve 5.0\% improved Acc@25. It is likely that the ground truth information is not complete, and thus for some methods, the additional information afforded by Mask3D predictions can provide greater signal. Furthermore, there may be a significant degree of sensitivity of these methods to the objects presented to the models.

  \textbf{ScanNet vs. ScanNet++}. We report results on ScanNet++ annotations in \autoref{tab:results-scannetpp}. ZSVG3D achieved as good or better performance on ScanNet++, in contrast with 3D-VisTA and 3D-GRAND. While the additional point cloud resolution may be useful for certain prompts, in practice the scenes are out of distribution and much larger, in terms of floorplan size, object counts, and semantic classes. While 3D-GRAND maxes out its input token limit, ZSVG3D scales at a slower rate to larger scenes and can still process all of the objects, albeit more slowly. Overall, grounding targets evidently becomes more difficult with many potential objects, necessitating future work to improve performance in challenging scenes.

  \textbf{Training dataset vs. performance}. All of the methods trained on LLM-scaled datasets, notably 3D-VisTA and 3D-GRAND, significantly outperformed the CLIP-based methods, which have difficulty parsing complex language relations \cite{yuksekgonul2023and}. However, PQ3D, trained on an aggregate dataset of majorly manually annotated prompts, outperformed the other methods, suggesting that simply scaling the volume of 3D-language pair data may not be a guarantee of better performance. Future work should explore these differences further to identify the effects of data vs. architecture on performance.

  \textbf{Subgroup analysis}. A detailed breakdown of key results is in \autoref{tab:result-subgroup}. On ground truth boxes, PQ3D achieves the best performance in nearly all categories, with ZSVG3D scoring better on prompts with text labels, ordinal relationships, comparisons, and agent-based anchors. However, these trends are volatile depending on whether GT or predicted boxes are provided (see \autoref{tab:result-subgroup-mask3d} for Mask3D subgroup performance).
  We highlight the following deficiencies in existing models:

  \textbf{Generic and coarse-grained target references}. Most models have lower performance on prompts without a fine-grained object class. PQ3D achieved comparable performance across all types of target references for both GT and Mask3D boxes.

  \textbf{Challenging attribute types}. Text labels were challenging for all models, due to insufficient point cloud resolution and inability of most methods to ingest RGB-D image data directly. Aside from ZSVG3D and PQ3D, models performed worse on number and state attributes as well.

  \textbf{Challenging relationship types}. ``Far'' and ``arrangement'' relationships were challenging for all models except PQ3D, while ordinal relationships were challenging for all models except ZSVG3D.

  \textbf{Negation}. Prompts with negation to describe attributes or relationships in most cases led to worse performance compared to those without. 3D-GRAND uniquely achieved relatively strong performance on negative prompts, with both ground truth and Mask3D predictions.

  The consistent drop in performance on \oursshort across all models compared to the existing benchmarks demonstrates a need for further improvement of models to achieve strong performance across a more diverse range of 3DVG prompts. While we do find that at least one model performs comparably on every subgroup compared to the control, no model consistently outperforms the others on all categories. Future work could bridge the gap in certain language phenomena for specific models or combine learnings across all models.


%% file: table/result_subgroup.tex
\begin{table*}[tb]
\centering
\resizebox{\linewidth}{!}
{
\begin{tabular}{@{}l rrrrrrrrrrrrrrrrr @{}}
\toprule
& & \multicolumn{3}{c}{Attributes} & \multicolumn{4}{c}{Relationships} & \multicolumn{4}{c}{Target Reference} & \multicolumn{4}{c}{Anchor Type} & \multicolumn{1}{c}{Lang} \\
\cmidrule(){3-5} \cmidrule(l){6-9} \cmidrule(l){10-13} \cmidrule(l){14-17} \cmidrule(l){18-18}
& Overall & Num & Lab & State & Far & Arr & Ord & Comp & Gen & CG & FG & NFN & Sing & Mul & Non & Agt & Neg \\
\midrule
OpenScene & 2.1 & 4.4 & 4.0 & 0.0 & 0.0 & 0.0 & 0.0 & 0.0 & 2.5 & 1.9 & 2.0 & 0.0 & 3.8 & 1.1 & 1.6 & 0.0 & 8.1 \\
LERF & 2.5 & 0.0 & 4.0 & 4.0 & 3.3 & 2.9 & 3.7 & 6.1 & 2.5 & 1.9 & 2.7 & 2.9 & 0.8 & 4.4 & 6.6 & 3.7 & 0.0 \\
ZSVG3D & 18.9 & 20.5 & \best{12.0} & \best{28.0} & 13.3 & 8.8 & \best{19.2} & \best{25.0} & 15.8 & 13.2 & 21.8 & 19.4 & 19.4 & 15.7 & 14.8 & \best{23.1} & 10.8 \\
LLM-Grounder & 2.5 & 2.2 & 0.0 & 0.0 & 3.3 & 5.7 & 11.1 & 6.1 & 0.0 & 0.0 & 4.1 & 7.2 & 1.5 & 5.5 & 4.9 & 11.1 & 2.7 \\
\midrule
3D-VisTA & 14.2 & 6.7 & 0.0 & 8.0 & 10.0 & 5.7 & 7.4 & 8.2 & 0.0 & 13.2 & 18.4 & 8.8 & 13.7 & 12.1 & 15.0 & 19.2 & 8.1 \\
3D-GRAND & 17.9 & 13.3 & 4.0 & 12.0 & 13.3 & 8.6 & 14.8 & 18.4 & 7.5 & 13.2 & 22.4 & 17.4 & 18.3 & 15.4 & 19.7 & 18.5 & \best{21.6} \\
PQ3D & \best{26.2} & \best{28.9} & 8.0 & \best{28.0} & \best{26.7} & \best{22.9} & 7.4 & 24.5 & \best{20.0} & \best{24.5} & \best{28.6} & \best{26.1} & \best{23.7} & \best{22.0} & \best{24.6} & 18.5 & 13.5 \\
\bottomrule
\end{tabular}
}
\caption{\textbf{Subgroup Analysis}. Breakdown of accuracy using ground truth boxes on \oursshort for ScanNet scenes across several subgroups of prompts. In general, we find that no one model is consistently better than another on any particular subgroup, likely suggesting that all of these models requires significant improvement to achieve any real understanding of the different linguistic phenomena and how they relate to 3D scenes.
}
\label{tab:result-subgroup}
\end{table*}


%% file: table/result_scannetpp.tex
\begin{table}[tb]
\centering
{
\begin{tabular}{@{}l rrrr @{}}
\toprule
& \multicolumn{2}{c}{ScanNet} & \multicolumn{2}{c}{ScanNet++} \\
\cmidrule(){2-3} \cmidrule(l){4-5}
& Acc & F1 & Acc & F1 \\
\midrule
ZSVG3D & \textbf{18.9} & 12.2 & \textbf{18.3} & \textbf{24.5} \\
3D-VisTA & 14.2 & 14.1 & 11.8 & 11.1 \\
3D-GRAND & 17.9 & \textbf{17.9} & 9.2 & 9.2 \\
\bottomrule
\end{tabular}
}
\caption{Accuracy and F1 score (\%) on \oursshort for ScanNet++ scenes, using ground truth boxes.
}
\label{tab:results-scannetpp}
\end{table}


%% file: figs/examples.tex
  \begin{figure*}[t]
  \centering
 \vspace{-0.4cm}
  \includegraphics[trim={0 4px 0 0},clip,width=\linewidth]{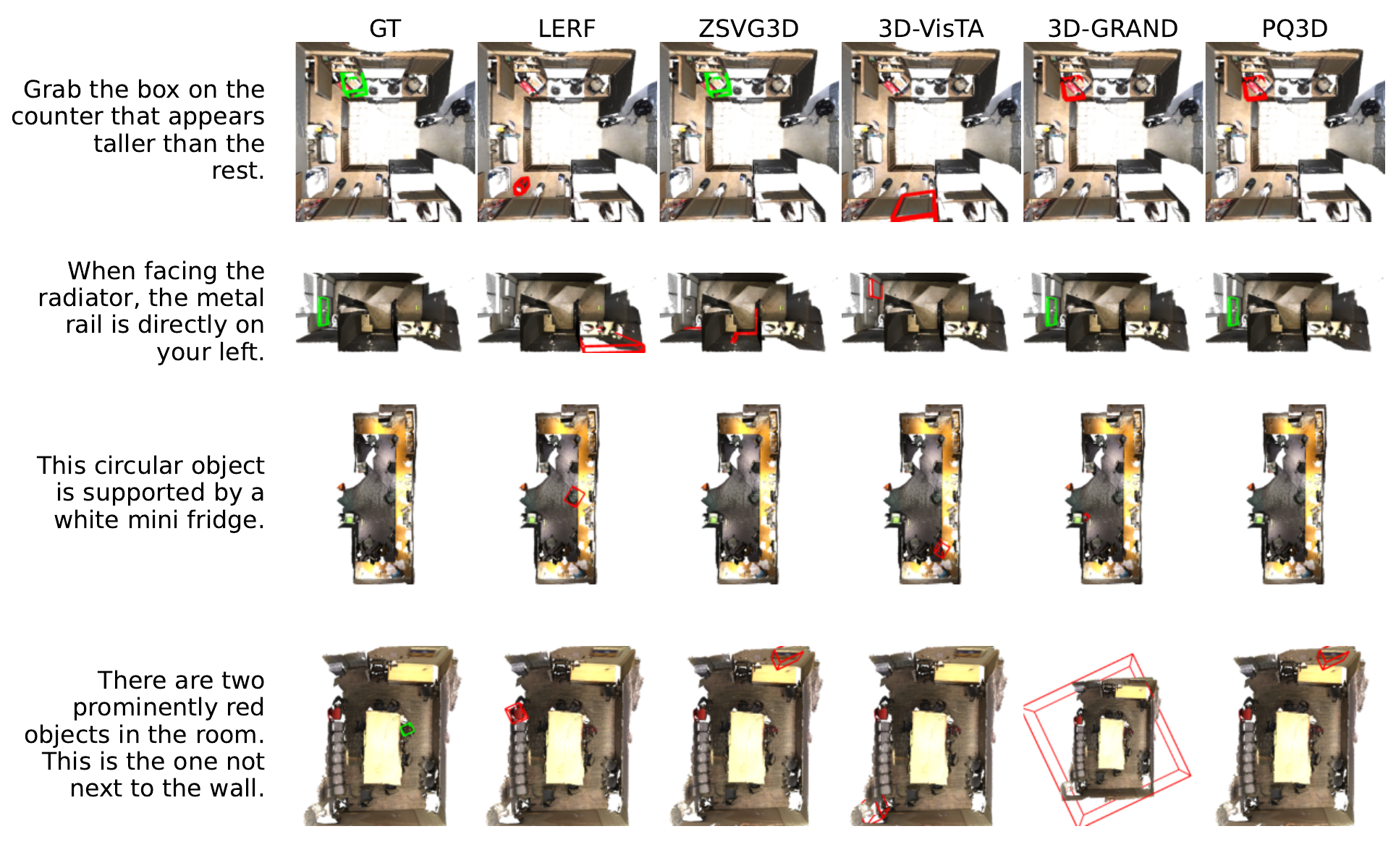}
  \vspace{-8mm}
  \caption{\textbf{Examples.} We show example predictions for each model on prompts with different linguistic patterns. 
  }
  \label{fig:examples}
 \vspace{-2mm}
\end{figure*}

%% file: section/paper/06_conclusion.tex
\section{Conclusion}
  \oursshort demonstrates the need for incorporating greater linguistic diversity when training and evaluating 3D visual grounding. We provide a framework and automated pipeline for analyzing language patterns and \oursshort to evaluate the successful parsing of different language patterns. Our analysis shows the need for further establishment of a comprehensive benchmark prompt and a need for better 3DVG performance on several subgroups of prompts.
  
  Scaling up \oursshort would be valuable in future development for large-scale training and evaluation, with an emphasis on more precise conditioning for language generation. While we have expanded the domain of visual grounding within language patterns, further work is also required to fully capture the complete space of potential prompts in the visual domain as well, toward ultimately utilizing learnings from 3DVG in general visual question-answering, embodied AI, and other applications.

%% file: section/paper/07_limitations.tex
\section{Limitations}

  We provide detailed and high quality annotations for evaluating visual grounding on 3D scenes. However, our dataset is relatively small, which may affect the power of conclusions one can draw from it. \vlmlimitations{Vision-language models (VLMs) are a natural solution for scaling, given their alleged flexibility in language and ability to parse visual inputs directly, but current state-of-the-art VLMs suffer from several key limitations.}
  
  Firstly, VLMs lack comprehensive 3D understanding. While they can identify basic spatial relationships, mapping objects across frames and extrapolating different viewpoints is challenging. 
  Furthermore, identifying multi-object relationships, such as ordinal positions of chairs in a lecture hall, is likewise difficult.
  \cameraready{When generating grounding descriptions from images and extracted captions, VLMs did not consistently reconcile information correctly across views or captions. Furthermore, VLMs rarely generated correct viewpoint-dependent prompts, suggesting an inability to ground objects from particular perspectives without significant guidance.
  }
  
  Secondly, \cameraready{VLMs do not inherently generate diverse descriptions, relying on in-context examples to condition the distribution~\citep{chang2024partnr}. Without a wide distribution of examples, VLM-generated descriptions tend to overfit to the provided examples. Even with stricter prompting and many in-context examples, VLMs in our trials did not necessarily capture the full distribution, for instance neglecting certain patterns or lacking variance in sentence length. Furthermore, prompting for increased diversity risked generating descriptions that are incorrect.}
  
  \cameraready{Lastly, they do not reliably reason about all of the objects in a scene---while they can caption single objects, grounding requires \emph{contrasting} objects and thus identifying whether attributes are also true of all other objects in the scene. We found that grounding descriptions generated directly from scene graphs using VLMs were usually true of the targets but failed to consistently differentiate them from other objects in the scene, even with the full scene context.}
  We believe that even at the modest scale of \oursshort, our evaluation, framework, and insights are still beneficial for the language and vision communities. Future work to address the limitations of VLMs will be key for scaling up the dataset in future work.

  While our prompts are focused on describing the contents of 3D scenes, potentially offensive language could be present. We focused primarily on physical descriptions of objects and their everyday use cases, and we reviewed our prompts to ensure that no such language was used.

  We focus on grounding objects in English. We acknowledge that different cultures may have different types of indoor scenes from those of ScanNet and ScanNet++ or different ways of describing objects. Future work should explore these differences further.

%% file: section/paper/09_acknowledgements.tex
\section*{Acknowledgement}
\label{sec:acknowledgement}
This work was funded in part by a CIFAR AI Chair and the NSERC Discovery Grant. 
Infrastructure was supported by CFI JELF grants.
We thank Yiming Zhang, Hou In Ivan Tam, Hou In Derek Pun, Xingguang Yan, and Karen Yeh for helpful discussions and feedback.

%% file: section/paper/08_supplement.tex
\appendix

\section*{Appendices}

We provide additional details about the analysis of prior datasets (\appref{appendix:dataset-analysis}, construction of ViGiL3D (\appref{appendix:dataset}),  implementation details for methods we compared (\appref{appendix:baselines}), and additional experiment results (\appref{appendix:results}).

\section{Dataset Analysis}
  \label{appendix:dataset-analysis}
  \input{table/dataset_overview}

  In this work, we focused on analyzing 3DVG datasets in English. \autoref{tab:dataset-overview} provides further details about each of the dataset we analyzed, including the statistics on the text description (e.g. prompts), the annotation method, and the source scene datasets.

  We document below the process for creating the language criteria used to evaluate datasets and model predictions as well as further details concerning the manual validation.

  \subsection{Criteria Selection}

  Semantic scene graphs, which can be used to represent the key objects in an image or scene and their attributes and relationships to one another, comprised the initial basis for criteria selection. While they are typically used to describe an entire scene, the concept also applies usefully to visual grounding, in which there is a special object or set of objects (\textit{target}) to identify. Thus the high-level categories for characterization included 1) how objects, especially the target, were specified; 2) the types of attributes; 3) the types of relationships; and 4) the grammatical structure that could be used to translate a scene graph to natural language.
  
  Given this framework, the specific criteria used to analyze existing 3DVG datasets and construct subgroups for analysis were selected based on qualitative observation of a sample of prompts from existing datasets. Based on observed similarities and notable gaps, we constructed a set of criteria (see \autoref{tab:dataset-criteria}) that could be quantitatively measured and thus reasonably executed by LLMs and natural language libraries.
  
  While there may be other, more comprehensive taxonomies by which to characterize grounding prompts, we believe that our system is 1) sufficiently detailed to identify the categories of patterns which could affect a model's ability to ground objects and 2) useful for characterizing the vast majority of 3DVG prompts in existing datasets and usefully highlights language patterns which are still under-represented.

  \subsection{Analysis Pipeline}
  \input{figs/dataset_radar_graph}
  \input{figs/object_prompt}
  \input{figs/relationship_prompt}
  \input{figs/attribute_prompt}

  In our pipeline, we use \texttt{gpt-4o-2024-08-06} \cite{achiam2023gpt} as our LLM to parse the grounding prompt into an augmented scene graph-like representation, as well as to extract certain linguistic properties of the prompt. We use the \texttt{en\_core\_web\_md} spaCy pipeline \cite{spacy2} for token and PoS parsing, including identifying negation and computing the unique bigram frequency.

  We include the three prompts used to automate the analysis of objects, relationships, and attributes in the prompts in Listings \ref{fig:object-prompt}, \ref{fig:relationship-prompt}, and \ref{fig:attribute-prompt}, respectively. The script, including OpenAI API calls, evaluates a single prompt in around 9 seconds per iteration, thus requiring a total of 21 compute-hours to generate the full analysis across all datasets, at a cost of around 23 USD per 1000 prompts. The burden of computation was primarily on GPT-4o \cite{achiam2023gpt}, so only CPU resources were required here.

  \subsection{Threshold Selection}
  \label{appendix:threshold-selection}

  \thresholdselect{We use thresholds of 5\% and 20\% to measure whether a particular prompt characteristic was sufficiently reflected in the aggregate performance, while accounting for the fact that not every prompt should reflect every characteristic. While many language patterns can co-occur, too much co-occurrence would prevent us from being able to test a model’s ability to parse particular patterns and allow models to over-attend to one phenomena at the exclusion of others. This is precisely what we observe in 3D-GRAND and ScanScribe, which use excessive attributes and relationships per prompt. Thus, there is a limit to the value of having a high proportion of prompts with a particular characteristic.}
  
  \thresholdselect{We found during \oursshort annotation that 20\% was a natural threshold, given that increasing the proportion of one pattern tends to reduce the proportions of others. Because most papers still ultimately compare performance on aggregate statistics, we opt to use target percentages over absolute thresholds on prompt counts per language pattern.}

  \subsection{Manual Validation}

  To manually validate the pipeline, a random sample of 20 prompts were annotated by the authors from each of the prior datasets as well as 100 prompts from \oursshort (total of 225 prompts). The quantitative design of the criteria made it straightforward to annotate each count or binary flag, and a similar prompt of examples were used as a guideline for each metric. For each of the 24 binary metrics, we computed a precision, recall, and F1 score, and for each of the counts (e.g., number of attributes per prompt), we measured the deviation using the median error, median absolute deviation, and mean absolute error. While a couple metrics with lower representation, such as the style attribute, were harder to predict correctly by the model, most of the criteria were predicted correctly by the model with above 80\% precision and recall. For most prompts, examples where the model predicted ``incorrectly'' were gray-area cases, suggesting that the model largely picked up on the right signals.

  \subsection{Additional Analysis}
  \label{appendix:additional-dataset-analysis}
  \input{table/dataset_comparison_all_attributes}
  \input{table/dataset_comparison_all_relationships}
  \input{table/dataset_comparison_all_others}

  \cameraready{We include the full analysis of linguistic properties in \cref{tab:dataset-comparison-attributes,tab:dataset-comparison-relationships,tab:dataset-comparison-other}. \autoref{tab:dataset-comparison-attributes} includes all of the attribute types analyzed by the pipeline, and \autoref{tab:dataset-comparison-relationships} includes all of the relationship types. \autoref{tab:dataset-comparison-other} contains all other analyzed metrics, including those for target reference, anchor type, and diversity. We see that unlike other datasets, \oursshort includes all of the linguistic patterns, with the highest prevalence in most categories. \oursshort furthermore is more balanced across linguistic properties compared to existing datasets.}

  While most datasets have parseable grounding prompts, some are invalid or confusing. 3D-VisTA has the largest number of parsing failures, including 6\% of prompts which could not be interpreted by the analysis pipeline. This is largely due to a lack of data cleaning, as a significant portion of the LLM-generated responses involve the LLM expressing inability to respond to the query (e.g., ``I'm sorry, I don't have enough information to answer that question...''). 

  Additionally, some attributes were reasonably identified by the pipeline as present in certain datasets despite not contributing meaningfully to grounding. For instance, SceneVerse has many prompts with embellishments which the pipeline interprets as ``state'' attributes (e.g., ``The bag rests gracefully on the floor's surface.''). However, while interesting in their own right and arguably not an error in parsing, it does not describe a useful state for grounding. Clear cases like these, when egregious, were manually corrected in postprocessing.

\clearpage
\clearpage

\section{\oursshort Dataset}
  \label{appendix:dataset}

  We provide further details here for the process to develop \oursshort.

  \subsection{Grounding Annotation}

  Prompts for \oursshort were annotated internally by the authors. The annotator demographic was primarily Asian and included a native English speaker. 
  Scenes were sampled from ScanNet \cite{dai2017scannet} and ScanNet++ \cite{yeshwanth2023scannet++}. ScanNet has 1513 scenes with RGB-D video and point cloud representations, reconstructed from real indoor scenes. ScanNet++ has 460 reconstructed scenes with similar RGB-D streams as well as DSLR images and laser scan data. While having fewer scenes, ScanNet++ has higher quality reconstructions and larger scenes overall. in order, primarily from the validation splits of ScanNet and ScanNet++, in order to minimize the possibility of other models having trained on the scenes we used. Both datasets redact any potentially identifying information in the videos and scenes, and our text prompts focus only on describing the contents of the scenes themselves.
  
  Annotators were provided a 3D point cloud view of each scene with access to a ground truth instance segmentation of the scene and an RGB video of an agent navigating the scene. For each prompt, they were instructed to select an object and then to craft a description identifying that object using a combination of the class name, attributes, or relationships to other entities in the scenes. In order to achieve linguistic diversity, annotators were instructed to use different sampled linguistic patterns for each prompt, in order to represent each phenomena in the dataset well. Each prompt was further annotated with metadata concerning the linguistic patterns present. Annotations were manually validated to ensure correctness of the target objects with respect to the descriptions and metadata.

  \subsection{Prompt Validation}

  \promptvalidation{We ran a study with human evaluators, presenting them with the 3D point cloud and corresponding RGB-D video of each scene and asking them to identify the target objects (0, 1, or multiple) of each grounding prompt. When presented with the same ground truth object segmentations as the models, we found that they achieved an overall accuracy of 84\% on the ScanNet prompts, significantly exceeding model performance. The main challenge encountered was searching through large scenes and videos, especially given their low resolution. This demonstrates that the prompts are sufficiently solvable and that existing models are not yet attaining human performance.}

\input{table/method_comparison}
\input{figs/model_diagram}
\section{Evaluation}
  \subsection{Implementation Details}
  \label{appendix:baselines}

  We describe the details of reproducing each method below, toward faithfully representing each model while also ensuring a fair comparison between them:

  \textbf{OpenScene}~\cite{peng2023openscene} is a simple contrastive learning-based approach which aligns 3D point features to projected 2D segmentation features in the CLIP space. We use the pretrained OpenSeg version of the model with \texttt{MinkUNet18A} for 3D encoding and evaluate using only the 3D features. To adapt OpenScene with provided boxes (from ground truth or Mask3D), we compute the mean cosine similarity of all embeddings for points contained within each box. We select the target object which has the highest mean similarity score.

  \textbf{LERF}~\cite{kerr2023lerf} is a model which augments neural radiance fields by learning a CLIP feature for each point. We use the LERF model based on ViT-B/16, using the \texttt{Nerfstudio} package \cite{nerfstudio}. We optimize a LERF model on each scene using every 20th frame from the RGB-D videos, at a frame resolution of $320\times 240$. We sample $10\times{6}$ points per scene for inference and compute the target object similarly to OpenScene.
  Note that as LERF requires optimizing the 3D representation for each scene, it is the most computationally expensive during inference.

  \textbf{ZSVG3D}~\cite{yuan2024visual} is an LLM-based method which leverages visual program synthesis for reasoning. We use GPT-4o \cite{achiam2023gpt} as the LLM to bring the method closer in line with the other methods, which use GPT-4 or variants. Furthermore, we achieve better results than GPT-3.5 as used in the original implementation, with only a 16.2\% accuracy on ground truth boxes with GPT-3.5 compared to 22.9\% with GPT-4o.
  The LOC module applies CLIP (ViT-B/16) to the ground truth or predicted labels of each box to compute alignment against the object of interest.

  \textbf{LLM-Grounder}~\cite{yang2024llm} is a zero-shot method using CLIP-based methods to detect objects and an LLM (GPT-4, as in the original implementation) to plan and reason to identify the target objects based on attributes and anchors. We use OpenScene as the visual grounder in our experiments. We observe that the performance was significantly decreased when using ground truth or Mask3D boxes compared to the original method of clustering using DBSCAN~\cite{ester1996density}. This is likely caused by the fact that the clustering is based on only those points whose OpenScene embeddings have a high cosine similarity with the text embeddings, as opposed to purely geometric clustering. Per the original implementation, the box dimensions provided to the LLM are based on the clustered points in the original method, causing the LLM to receive different box dimensions than those of the actual object. Interestingly, we find that modifying the implementation to serve the original box dimensions to GPT instead of using the cluster-extract ones yielded slightly worse performance.

  \textbf{3D-VisTA}~\cite{yang20243d} is pretrained on the ScanScribe dataset and uses a simplified dual-encoder transformer architecture to perform a variety of downstream 3D tasks. We use the pretrained checkpoint from the authors fine-tuned on ScanRefer to optimize performance for visual grounding on ScanNet-like scenes. In order to support larger scenes, the point cloud sequence length was extended from 80 to 250.

  \textbf{3D-GRAND}~\cite{yang20243d} is trained on the dataset of its namesake, representing 3D-LLM performance on a significantly larger dataset of synthetic scenes. We use the 
\texttt{merged\_weights\_grounded\_obj\_ref} model checkpoint based on Llama-2-7b. Due to the large number of Mask3D predictions and 2048 input sequence token limit, we truncate the input object tokens accordingly.

  \textbf{PQ3D}~\cite{zhu2024unifying} is a promptable query transformer-based model for 3D, unifying different 3D representations, modality prompts, and output forms through prompting. The model is pretrained on 8 different 3D datasets, including notably ScanRefer, Nr3D, Sr3D, and Multi3DRefer. We use the provided checkpoint for the unified 3D model, after the second stage of training. The pretrained model based on CLIP (ViT-L/14) and uses PointNet++ for point cloud encoding.

  In general, all LLM-based methods were executed multiple times in cases where failures occurred for specific prompts during inference, largely due to syntactic errors in parsing the LLM output. ZSVG3D encountered the most errors due to unparseable outputs, such as calling functions in its generated programs which were not supported by the domain language, and we were unable to generate a valid output for around 2\% of prompts after 5 attempts.

  A summary of the models evaluated can be found in \autoref{tab:method-comparison}.

  \subsection{Additional Results}
  \label{appendix:results}

  \input{figs/model_radar_graph}

  \input{table/result_subgroup_mask3d}
  
\input{figs/examples_supp_scannet}
  \input{figs/examples_supp_scannetpp}

  We report the subgroup analysis based on Mask3D boxes in \autoref{tab:result-subgroup-mask3d}. As expected, the models that have the best aggregate performance, 3D-VisTA and 3D-GRAND, with Mask3D predictions achieve the best performance across most subgroups, while ZSVG3D and PQ3D are largely outperformed in most categories except prompts with \textit{state} attributes and \textit{generic} target references, respectively.

  We also provide additional qualitative examples of model performance on \oursshort with ScanNet and ScanNet++ scenes in Figures \ref{fig:examples-scannet} and \ref{fig:examples-scannetpp}, respectively.

  \statistical{\textbf{Statistical Analysis.} We find that, despite the dataset size, many of the results we observe in our subgroup analysis are statistically significant, such as the performance of ground truth vs. predicted boxes for PQ3D and ZSVG3D and the performance on generic target specification for 3D-VisTA. We use the 2-tailed two proportion z-test to compare the accuracies of the higher performing models on each subgroup compared to the performance on the rest of the ViGiL3D prompts for ScanNet scenes.}

  \reweightedeval{\textbf{Reweighted evaluation.} We analyze performance when reweighting the distribution of prompts to more closely match that of linguistic patterns in ScanRefer. This was to assess whether the drop in performance on \oursshort compared to ScanRefer is caused more by the presence or absence of specific patterns or another factor.
  We find that the model performance only improves marginally, such as for 3D-GRAND from 0.18 to 0.19 when reweighted to its training dataset. This is likely caused in part by the difficulty in estimating the frequency of every co-occurrence of 25 language patterns, in which a simplified model causes a regression toward the unweighted mean. We also hypothesize that our dataset is more challenging, even in the language patterns represented more prominently in other datasets, thus causing the increase to be fairly marginal: 
  1) the targets can be 0, 1, or multiple objects, whereas most datasets specify exactly one object; 
  2) \oursshort prompts are designed to specify attributes or relationships that must each be parsed successfully to identify the possible targets, whereas most other VG prompts refer to unique object classes or overspecify constraints; and 3) within each language pattern, there is still potential for large variation, as evidenced by \oursshort having the highest frequency of unique bigrams.
  Even if the model has seen a particular pattern before, its performance may still be poor as a result of these additional challenges, and we observe both cases where the model performed better and worse on subgroups which were frequent in their training set.}

\section{Scientific Artifacts}
The licenses used in this paper include the following: ScanNet (terms of use\footnote{\url{https://kaldir.vc.in.tum.de/scannet/ScanNet_TOS.pdf}}), ScanNet++ (terms of use\footnote{\url{https://kaldir.vc.in.tum.de/scannetpp/static/scannetpp-terms-of-use.pdf}}) ScanRefer (CC BY-NC-SA 3.0), Nr3D/Sr3D+ (MIT), Multi3DRefer (MIT), 3D-GRAND (CC BY 4.0), 3D-VisTA and ScanScribe (MIT), SceneVerse (terms of use \footnote{\url{https://drive.google.com/file/d/14Ji7PLOKsAxrXpxV6EWLsQGjzcEuk35N/view}}), Instruct3D (CC BY-NC-SA 4.0), OpenScene (apache-2.0), LERF (MIT), ZSVG3D (N/A), LLM-Grounder (apache-2.0), PQ3D (MIT), OpenAI (terms of use\footnote{\url{https://openai.com/policies/terms-of-use/}}), and spaCy (MIT). We follow the intended use of all of the licenses in the paper and reported our intended usage in the terms as appropriate.

LLMs, including ChatGPT, were used in the analysis pipeline and in some of the baseline methods. We also used them as assistive tools for generating code and researching methodologies.

%% file: table/dataset_overview.tex
\begin{table*}[tb]
\centering
\resizebox{\linewidth}{!}
{
\begin{tabular}{@{}lrrc p{2.5in} p{3.5in} @{}}
\toprule
Dataset & \# Prompts & $\abs{V}$ & Gen Method & Scene Datasets & Example \\
\midrule
ScanRefer & 52K & 6,919 & Human & ScanNet & {\small There is a black counter top to the left of the fridge. It has a stainless steel sink on it.} \\
Nr3D & 42K & 6,951 & Human & ScanNet & {\small While at the sink, it is the third option on the top.} \\
Sr3D+ & 115K & 200 & Template & ScanNet & {\small The whiteboard that is close to the couch} \\
Multi3DRefer & 62K & 9,645 & Human, LLM & ScanNet & {\small The black frame houses the picture, and it hangs above the toilet.} \\
3D-GRAND & 6.2M & 8,279 & Template, LLM & 3D-FRONT, Structured3D & {\small This refrigerator is a muted silver, presenting a sleek and modern look with its brushed metal finish. The refrigerator is positioned close to one of the dining chairs, near to another dining chair, and far from the loveseat sofa.} \\
ScanScribe & 278K & 2,881 & Human, Template, LLM & ScanNet, 3RScan & {\small The chair is behind the desk, on the left side of the circular black table, to the right of the rectangular shelf, and to the left of the other chair, bag, trash can, and backpack.} \\
SceneVerse & 2.5M & 18,427 & Template, LLM & ProcTHOR, Structured3D, ARKitScenes, HM3D, ScanNet, 3RScan, MultiScan & {\small The couch is situated to the right of the computer tower.} \\
Instruct3D & 2.6K & 1,787 & Human & ScanNet++ & {\small Where are the window coverings used to control light and privacy? It is the one faces two doors and black chair.} \\
\bottomrule
\end{tabular}
}
\caption{\textbf{Dataset Overview}. Overview of dataset sizes and example prompts from each prior dataset. We include the size of each dataset as represented by the number of 3DVG prompts and vocabulary size ($\abs{V}$), as well as the underlying scene datasets. The sizes are calculated by counting distinct words across the dataset prompts used for visual grounding, which may be a subset of the total number of prompts provided in each dataset (particularly for 3D-GRAND, ScanScribe, and SceneVerse).
}
\label{tab:all-dataset-overview}
\end{table*}


%% file: figs/dataset_radar_graph.tex
\begin{figure}[ht]
  \centering
  \includegraphics[trim={0 0 0 5px},clip,width=\linewidth]{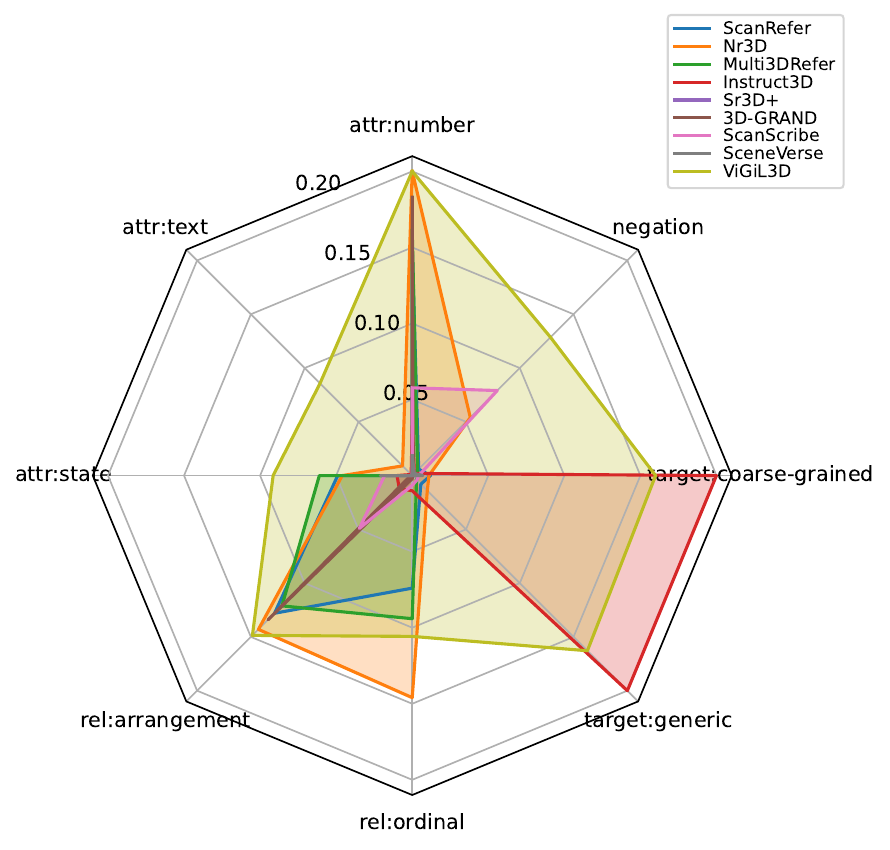}
\captionof{figure}{
  \textbf{Dataset Comparison.} We visualize the proportion of prompts representing different metrics up to 20\% of the dataset size.
}
  \label{fig:dataset-radar}
\end{figure}

%% file: figs/object_prompt.tex
\begin{listing*}[t]
  \centering
  \begin{lstlisting}[language=Python,frame=single,breaklines=true]
"""
You are given a description of an object that someone is supposed to find in a scene. Similar to the Visual Genome dataset, we would like to identify the objects, attributes, and relationships in the following text:
"{}"

Please first return a list of the objects in the scene in a JSON format:
{{
  "success": boolean,
  "objects": [
    {{
      "id": id of object,
      "name": name of object as string,
    }}
  ],
  "target": list of ids of object,
  "target_reference_type": "generic", "categorical", or "fine-grained",
  "first_noun": boolean
}}

The object IDs should start with 0 and increment.

The name of the object should be lowercase (except for proper nouns) and sufficient to define the object class, if specified. Attributes should not be included in the class name. For instance, "a big, red apple" has the object name of "apple", and a "rectangular washing machine" has the object name of "washing machine". Parts of objects should be included as separate objects. For instance, "a chair with four legs" has objects "chair" and "legs".

If the target object is implied in the text but not explicitly named, such as in, "Where can I keep food cold?" the target object name should just be specified as "object".

The target ID should be the list of IDs of the objects that the user is supposed to find. For example, if the prompt is, "This is the toolbox on the shelf", the target ID should be the ID of the toolbox.

target_reference_type should be specified as the most specific term used to name the object:
* generic - object is referred to by a general term such as "object", "thing", or "item", or if the object is implied but not explicitly named.
* categorical - object is referred to by a category which is more specific than a generic reference but not specific to a particular object class. This includes references such as "appliance", "seat", "container", "display", "machine", or "device".
* fine-grained - object is referred to by a specific object class, from which it should be easy to infer what the object is and how it should be used.

If the prompt is not a description of an object in a scene, set "success" to False and ignore the rest of the output. Otherwise, set it to True.

If the target object is the first noun phrase mentioned in the description, set "first_noun" to True. Otherwise, set it to False.
"""
  \end{lstlisting}
  \caption{\textbf{Object Prompt for Analysis.} We feed this prompt first to GPT-4o to obtain information about the objects in the description and their respective types.
  }
  \label{fig:object-prompt}
\end{listing*}

%% file: figs/relationship_prompt.tex
\begin{listing*}[t]
  \centering
  \begin{lstlisting}[language=Python,frame=single,breaklines=true]
"""
You are given a description of an object that someone is supposed to find in a scene. Your goal is to analyze the prompt for information about the relationships used to describe objects in the description:
"{}"
The parsed list of objects is as follows:
{}
Return the output in a JSON format according to the following format:
{{
  "relationships": [
    {{
        "name": name of relationship as string as it appears in the prompt,
        "subject_id": list of ids of objects which are the subject of the relationship,
        "recipient_id": list of ids of objects which is the recipient of the relationship
    }}
  ],
  "num_relationship_type": {{
    "near": {{
      "exists": True if relationship is found in prompt or False otherwise,
      "explanation": list of relationships identified, or empty if none
    }},
    "far": {{
      "exists": True if relationship is found in prompt or False otherwise,
      "explanation": list of relationships identified, or empty if none
    }},
    "viewpoint_dependent": {{
      "exists": True if relationship is found in prompt or False otherwise,
      "explanation": list of relationships identified, or empty if none
    }},
    "vertical": {{
      "exists": True if relationship is found in prompt or False otherwise,
      "explanation": list of relationships identified, or empty if none
    }},
    "contain": {{
      "exists": True if relationship is found in prompt or False otherwise,
      "explanation": list of relationships identified, or empty if none
    }},
    "arrangement": {{
      "exists": True if relationship is found in prompt or False otherwise,
      "explanation": list of relationships identified, or empty if none
    }},
    "ordinal": {{
      "exists": True if relationship is found in prompt or False otherwise,
      "explanation": list of relationships identified, or empty if none
    }},
    "comparison": {{
      "exists": True if relationship is found in prompt or False otherwise,
      "explanation": list of relationships identified, or empty if none
    }}
  }},
  "anchors": {{
    "single": True if at least one of the anchor objects is a single object otherwise False,
    "multiple": True if at least one of the anchor objects represents multiple objects otherwise False,
    "non_object": True if at least one of the anchor objects represents a region or room otherwise False
    "viewpoint": True if one of the relationships requires a specific viewpoint otherwise False
  }}
}}
A relationship compares an object(s) or region to another object(s) or region. Relationships should capture objects in the scene and not hypothetical objects. The name of the relationship should be the word or phrase used in the description to describe the relationship. If a noun in the description is a part of an object rather than a distinct object, then it should be counted as a part rather than an object. If a recipient is relative to the speaker of the description, use the ID of the "<speaker>" object.
Each relationship type is defined as follows:
near: Any relationship which describes one object in proximity of another. Examples include near, next to, nearby, adjacent, close to, proximate, amidst, among, covered, or contact relationships (against, leaning on, on, hanging on, supported by, attached to).
far: Any relationship which describes one object far away from another. Examples include far from, opposite, across from, and distant from.
viewpoint_dependent: Any relationship which can only be identified based on a canonical reference frame of the object or the viewpoint of the speaker. This includes left, right, in front of, facing, behind, or any cardinal direction.
vertical: Any relationship which describes one object above or below another. Examples include above, below, on top of, under, underneath, or vertical support relationships (e.g., an object on another).
contain: Any relationship which describes one object contained within another or some part that belongs to another. Examples include in, inside, within, with, has, or have.
arrangement: Any relationship which describes one object as part of an ordered arrangement. Examples include "between", "surrounded by", "row of", "column of", "stack of", or "pile of" other objects. You should exclude "amidst", "among", "nearby" or other non-structured relationships.
ordinal: Any relationship which describes the numerical position of an object in a spatial order or array. Examples include first, 2nd, middle, last. You should exclude cases of an object being the closest, leftmost, rightmost, or equivalent.
comparison: Any relationship which compares properties of different objects and requires identifying which one is more or less, or the most or least, of something. Examples include taller, tallest, shorter, greenest, closest, furthest, or same as.
In the explanation, each relationship should be given as a list of [subject, relationship, recipient].
Lastly, indicate the following:
1. If any of the subjects or recipients of a relationship, excluding the target, is a single object, set "single" to True
2. If any of the subjects or recipients of a relationship, excluding the target, represents multiple objects, set "multiple" to True. Examples include "the table is surrounded by six chairs", "the car is in between the shovel and the desk", "the book is the third one on the shelf", or "the chair is the one closest to the door".
3. If any of the subjects or recipients of a relationship is a region or room, set "non_object" to True. Examples include "the shelf in the center of the room", "the microwave in the kitchen", or "the books in the area around the couch".
4. If finding the target is dependent on a specific viewpoint in the scene, set "viewpoint" to True. Examples include "the leftmost wall" or "the window on your right."
"""
  \end{lstlisting}
 \vspace{-4mm}
  \caption{\textbf{Relationships Prompt for Analysis.} We then feed this prompt to GPT-4o, using the object information, to obtain information about the relationships between the inferred objects in the prompt.
  }
  \label{fig:relationship-prompt}
\end{listing*}

%% file: figs/attribute_prompt.tex
\begin{listing*}[t]
  \centering
  \begin{lstlisting}[language=Python,frame=single,breaklines=true]
"""
You are given a description of an object that someone is supposed to find in a scene. Your goal is to calculate statistics about the attributes used to describe objects in the description:
"{}"
The parsed list of objects is as follows:
{}
The parsed list of relationships is as follows:
{}
Return the output in a JSON format according to the following format:
{{
  "num_attribute_type": {{
    "color": {{
        "exists": True if attribute is found in prompt or False otherwise,
        "explanation": list of attributes identified, or empty if none
    }},
    "size": {{
        "exists": True if attribute is found in prompt or False otherwise,
        "explanation": list of attributes identified, or empty if none
    }},
    "shape": {{
        "exists": True if attribute is found in prompt or False otherwise,
        "explanation": list of attributes identified, or empty if none
    }},
    "number": {{
        "exists": True if attribute is found in prompt or False otherwise,
        "explanation": list of attributes identified, or empty if none
    }},
    "material": {{
        "exists": True if attribute is found in prompt or False otherwise,
        "explanation": list of attributes identified, or empty if none
    }},
    "texture": {{
        "exists": True if attribute is found in prompt or False otherwise,
        "explanation": list of attributes identified, or empty if none
    }},
    "function": {{
        "exists": True if attribute is found in prompt or False otherwise,
        "explanation": list of attributes identified, or empty if none
    }},
    "style": {{
        "exists": True if attribute is found in prompt or False otherwise,
        "explanation": list of attributes identified, or empty if none
    }},
    "text_label": {{
        "exists": True if attribute is found in prompt or False otherwise,
        "explanation": list of attributes identified, or empty if none
    }},
    "state": {{
        "exists": True if attribute is found in prompt or False otherwise,
        "explanation": list of attributes identified, or empty if none
    }}
  }},
  "attributes": [
    {{
      "object_id": id of object,
      "attributes": list of attributes
    }}
  ]
}}
Attributes are any descriptors that help distinguish an object from others. The name of an object does NOT count as an attribute.
Each attribute type is defined as follows:
color: Any attribute which describes the color properties of an object. Examples include red, blue, black, light, dark, monocolor, or colorful.
size: Any attribute which describes the size of an object. Examples include big, small, large, larger, tall, long, short, or medium. You should exclude cases where the height of an object is described to capture vertical position rather than size.
shape: Any attribute which describes the shape or form of an object. Examples include round, square, rectangular, or circular.
number: Any attribute which describes the quantity of an object. Examples include "two chairs". This does not include cases where the number is used to describe the relative order of the object, or cases where "one" is used as a pronoun to refer to the object.
material: Any attribute which describes what an object is made of. Examples include wood, metal, plastic, or glass. If the attribute describes the texture but not what the object is actually made of, e.g. metallic, then it should count as a texture attribute rather than a material.
texture: Any attribute which describes the texture of an object. Examples include smooth, rough, soft, metallic, or comfy.
function: Any attribute which describes what an object can be used for or the function it performs in a space. Examples include a chair for sitting or a lamp that makes the space warm and welcoming. The name of the object does not count as a function.
style: Any attribute which describes the style of an object or the effect of its presence in the space. Examples include modern, vintage, antique, futuristic, luxurious, or industrial, or describing its prominent or subtle presence in a room.
text_label: Any attribute which describes text that can be found on an object. Examples include "fragile" on a box or "exit" on a door.
state: Any attribute which describes the state of an object, which can be changed. Examples include "unopened" to describe a jar, "broken", or "drying" to describe clothes hanging on a rack.
You should also include a list of each of the attributes for each object in the scene.
"""
  \end{lstlisting}
 \vspace{-4mm}
  \caption{\textbf{Attributes Prompt for Analysis.} Finally, we feed this prompt to GPT-4o, substituting in the object and relationship information, to obtain information about the attributes in the prompt describing the inferred objects and their respective types. Including the relationship information helps the model to not double-count relationships as attributes.
  }
  \label{fig:attribute-prompt}
\end{listing*}

%% file: table/dataset_comparison_all_attributes.tex
\begin{table*}[tb]
\centering
\resizebox{\linewidth}{!}
{
\begin{tabular}{@{}l ccccccccccccc @{}}
\toprule
& All & Tgt & Anc & Col & Siz & Sha & Num & Mat & Fun & Tex & Sty & Lab & Sta \\
\midrule
ScanRefer & 1.90 & 1.21 & 0.68 & \doublecheckmark & \mycheckmark & \mycheckmark & \mycheckmark & \mycheckmark & \myxmark & \myxmark & \myxmark & \myxmark & \myxmark \\
Nr3D & 1.16 & 0.64 & 0.52 & \doublecheckmark & \mycheckmark & \myxmark & \doublecheckmark & \myxmark & \myxmark & \myxmark & \myxmark & \myxmark & \myxmark \\
Sr3D+ & 0.05 & 0.02 & 0.03 & \myxmark & \myxmark & \myxmark & \myxmark & \myxmark & \myxmark & \myxmark & \myxmark & \myxmark & \myxmark \\
Multi3DRefer & 1.73 & 1.21 & 0.52 & \doublecheckmark & \mycheckmark & \doublecheckmark & \mycheckmark & \mycheckmark & \myxmark & \myxmark & \myxmark & \myxmark & \mycheckmark \\
3D-GRAND & 5.81 & 4.68 & 1.12 & \doublecheckmark & \mycheckmark & \doublecheckmark & \mycheckmark & \doublecheckmark & \doublecheckmark & \doublecheckmark & \doublecheckmark & \myxmark & \myxmark \\
ScanScribe & 6.04 & 1.15 & 4.89 & \doublecheckmark & \myxmark & \doublecheckmark & \myxmark & \mycheckmark & \myxmark & \myxmark & \myxmark & \myxmark & \myxmark \\
SceneVerse & 0.41 & 0.35 & 0.07 & \myxmark & \mycheckmark & \myxmark & \myxmark & \myxmark & \myxmark & \mycheckmark & \mycheckmark & \myxmark & \myxmark \\
Instruct3D & 1.40 & 1.30 & 0.09 & \myxmark & \myxmark & \myxmark & \myxmark & \myxmark & \myxmark & \doublecheckmark & \myxmark & \myxmark & \myxmark \\
\rowcolor{gray!20} \oursshort & 1.62 & 1.09 & 0.53 & \doublecheckmark & \mycheckmark & \mycheckmark & \doublecheckmark & \mycheckmark & \mycheckmark & \mycheckmark & \mycheckmark & \mycheckmark & \mycheckmark \\
\bottomrule
\end{tabular}
}
\caption{\textbf{Dataset Attributes Comparison}. We show here a comparison of visual grounding datasets by attributes. The descriptions of each attribute type are provided in \autoref{tab:attributes}. Binary metrics for attribute types are reported as not present (\myxmark), some (\mycheckmark), or a lot (\doublecheckmark), as thresholded at 5\% and 20\% of the dataset sample.
}
\label{tab:dataset-comparison-attributes}
\end{table*}


%% file: table/dataset_comparison_all_relationships.tex
\begin{table*}[tb]
\centering
\resizebox{\linewidth}{!}
{
\begin{tabular}{@{}l ccccccccccc @{}}
\toprule
& All & Tgt & Anc & Near & Far & Dir & Ver & Cont & Arr & Ord & Comp \\
\midrule
ScanRefer & 2.33 & 1.89 & 0.44 & \doublecheckmark & \mycheckmark & \doublecheckmark & \doublecheckmark & \doublecheckmark & \mycheckmark & \mycheckmark & \myxmark \\
Nr3D & 2.22 & 1.63 & 0.59 & \doublecheckmark & \mycheckmark & \doublecheckmark & \doublecheckmark & \mycheckmark & \mycheckmark & \mycheckmark & \doublecheckmark \\
Sr3D+ & 1.00 & 1.00 & 0.00 & \doublecheckmark & \doublecheckmark & \myxmark & \myxmark & \myxmark & \myxmark & \myxmark & \doublecheckmark \\
Multi3DRefer & 2.02 & 1.63 & 0.39 & \doublecheckmark & \mycheckmark & \doublecheckmark & \doublecheckmark & \doublecheckmark & \mycheckmark & \mycheckmark & \myxmark \\
3D-GRAND & 2.81 & 2.71 & 0.10 & \doublecheckmark & \mycheckmark & \doublecheckmark & \doublecheckmark & \doublecheckmark & \mycheckmark & \myxmark & \mycheckmark \\
ScanScribe & 3.55 & 3.35 & 0.21 & \doublecheckmark & \myxmark & \doublecheckmark & \doublecheckmark & \mycheckmark & \myxmark & \myxmark & \doublecheckmark \\
SceneVerse & 1.33 & 1.30 & 0.03 & \doublecheckmark & \mycheckmark & \doublecheckmark & \doublecheckmark & \mycheckmark & \myxmark & \myxmark & \myxmark \\
Instruct3D & 1.43 & 1.31 & 0.12 & \doublecheckmark & \myxmark & \myxmark & \doublecheckmark & \doublecheckmark & \myxmark & \myxmark & \myxmark \\
\rowcolor{gray!20} \oursshort & 1.82 & 1.46 & 0.35 & \doublecheckmark & \mycheckmark & \doublecheckmark & \doublecheckmark & \doublecheckmark & \mycheckmark & \mycheckmark & \doublecheckmark \\
\bottomrule
\end{tabular}
}
\caption{\textbf{Dataset Relationships Comparison}. We show here a comparison of visual grounding datasets by relationships. The descriptions of each relationship type are provided in \autoref{tab:spatial-relationships}. Binary metrics for relationships types are reported as not present (\myxmark), some (\mycheckmark), or a lot (\doublecheckmark), as thresholded at 5\% and 20\% of the dataset sample.
}
\label{tab:dataset-comparison-relationships}
\end{table*}


%% file: table/dataset_comparison_all_others.tex
\begin{table*}[tb]
\centering
\resizebox{\linewidth}{!}
{
\begin{tabular}{@{}l ccccccccccc @{}}
\toprule
& Gen & CG & FG & NFN & Cor & Sing & Mul & Non & Agt & Neg & 2lex \\
\midrule
ScanRefer & \myxmark & \myxmark & \doublecheckmark & \myxmark & \doublecheckmark & \doublecheckmark & \mycheckmark & \doublecheckmark & \doublecheckmark & \myxmark & 0.20 \\
Nr3D & \myxmark & \myxmark & \doublecheckmark & \doublecheckmark & \mycheckmark & \doublecheckmark & \doublecheckmark & \doublecheckmark & \doublecheckmark & \mycheckmark & 0.27 \\
Sr3D+ & \myxmark & \myxmark & \doublecheckmark & \myxmark & \myxmark & \doublecheckmark & \myxmark & \myxmark & \myxmark & \myxmark & 0.02 \\
Multi3DRefer & \myxmark & \myxmark & \doublecheckmark & \doublecheckmark & \mycheckmark & \doublecheckmark & \mycheckmark & \doublecheckmark & \doublecheckmark & \myxmark & 0.28 \\
3D-GRAND & \myxmark & \myxmark & \doublecheckmark & \myxmark & \doublecheckmark & \doublecheckmark & \doublecheckmark & \doublecheckmark & \doublecheckmark & \myxmark & 0.05 \\
ScanScribe & \myxmark & \myxmark & \doublecheckmark & \myxmark & \doublecheckmark & \doublecheckmark & \doublecheckmark & \mycheckmark & \doublecheckmark & \myxmark & 0.10 \\
SceneVerse & \myxmark & \myxmark & \doublecheckmark & \mycheckmark & \mycheckmark & \doublecheckmark & \mycheckmark & \myxmark & \myxmark & \myxmark & 0.16 \\
Instruct3D & \doublecheckmark & \doublecheckmark & \mycheckmark & \doublecheckmark & \myxmark & \doublecheckmark & \myxmark & \doublecheckmark & \myxmark & \myxmark & 0.27 \\
\rowcolor{gray!20} \oursshort & \mycheckmark & \mycheckmark & \doublecheckmark & \doublecheckmark & \doublecheckmark & \doublecheckmark & \doublecheckmark & \doublecheckmark & \doublecheckmark & \mycheckmark & 0.45 \\
\bottomrule
\end{tabular}
}
\caption{\textbf{Additional Dataset Comparison}. We show here a comparison of visual grounding datasets by metrics for target reference, anchor type, and diversity. The descriptions of each metric are provided in \autoref{tab:dataset-criteria}. Binary metrics are reported as not present (\myxmark), some (\mycheckmark), or a lot (\doublecheckmark), as thresholded at 5\% and 20\% of the dataset sample.
}
\label{tab:dataset-comparison-other}
\end{table*}


%% file: table/method_comparison.tex

\begin{table*}[tb]
\centering
\resizebox{\linewidth}{!}{
\begin{tabular}{@{}l rccccr@{}}
\toprule
& \# Parameters & Visual Encoder & Text Encoder & Training Strategy & Training Dataset & Inference Time \\
\midrule
OpenScene & 100M & MinkUNet18A & CLIP text encoder & CLIP aligned & ScanNet200 & 0.12 \\
LERF & 237M & CLIP-ViT & CLIP text encoder & CLIP aligned & ScanNet & 23m \\
ZSVG3D & 149M\textsuperscript{*} & CLIP-ViT & GPT-4o & Zero-shot & N/A & 3.81 \\
LLM-Grounder & 100M\textsuperscript{*} & CLIP-ViT & GPT-4 & Zero-shot & ScanNet & 40.70 \\
3D-VisTA & 138M & PointNet++ & BERT & 3DVG data & ScanScribe & 0.02 \\
3D-GRAND & 6.74B & Mask3D & Llama-2 & 3DVG data & 3D-GRAND & 2.31 \\
PQ3D & 248M & CLIP-ViT,PointNet++  & CLIP text encoder & 3DVG data & Aggregate & 0.31 \\
\bottomrule
\end{tabular}
}
\caption{\textbf{Method Comparison}. Overview of the 3D visual grounding methods. We present the visual and text encoder they use, along with their training strategies and datasets. PQ3D is trained on an aggregate dataset including ScanRefer, Nr3D, Sr3D, Multi3DRefer, and other segmentation, QA, and captioning datasets. Inference times are reported in seconds, unless otherwise specified, on ScanNet with a batch size of 1. \textsuperscript{*}The number of parameters of GPT is not included, since it has not been publicly reported.
}
\label{tab:method-comparison}
\end{table*}

%% file: figs/model_diagram.tex
\begin{figure*}[tb]
  \centering
  \includegraphics[trim={0 0 0 5px},clip,width=\linewidth]{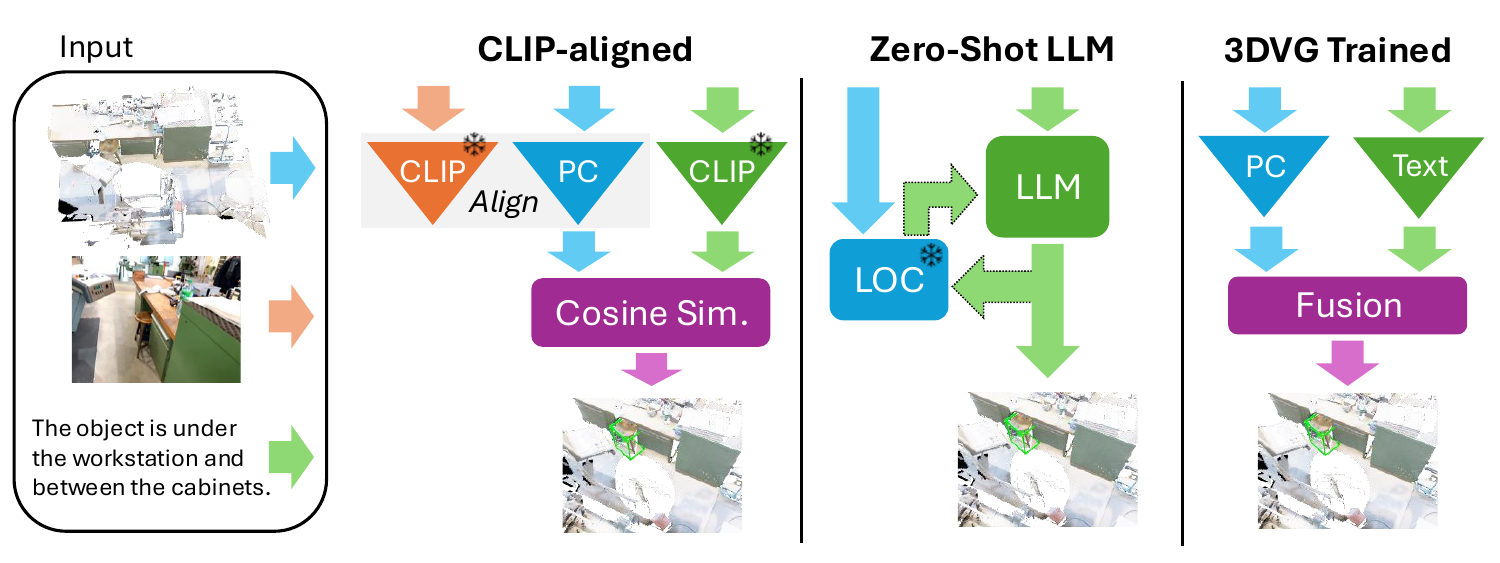}
\captionof{figure}{
  \textbf{3DVG Model Architectures.} \emph{Clip-aligned} models align a 3D point cloud encoder with a pretrained CLIP image encoder during training, resulting in an alignment between the point encoder and CLIP text encoder. \emph{Zero-shot LLMs} are used as language understanding and reasoning modules, employing a pretrained localization module to identify objects by class in the scene. \emph{3DVG-trained} models are trained with supervision on scaled 3D datasets, relying on stronger fusion to correspond embeddings.
}
  \label{fig:model-diagram}
\end{figure*}

%% file: figs/model_radar_graph.tex
\begin{figure}[htb]
  \centering
  \includegraphics[trim={0 0 0 5px},clip,width=\linewidth]{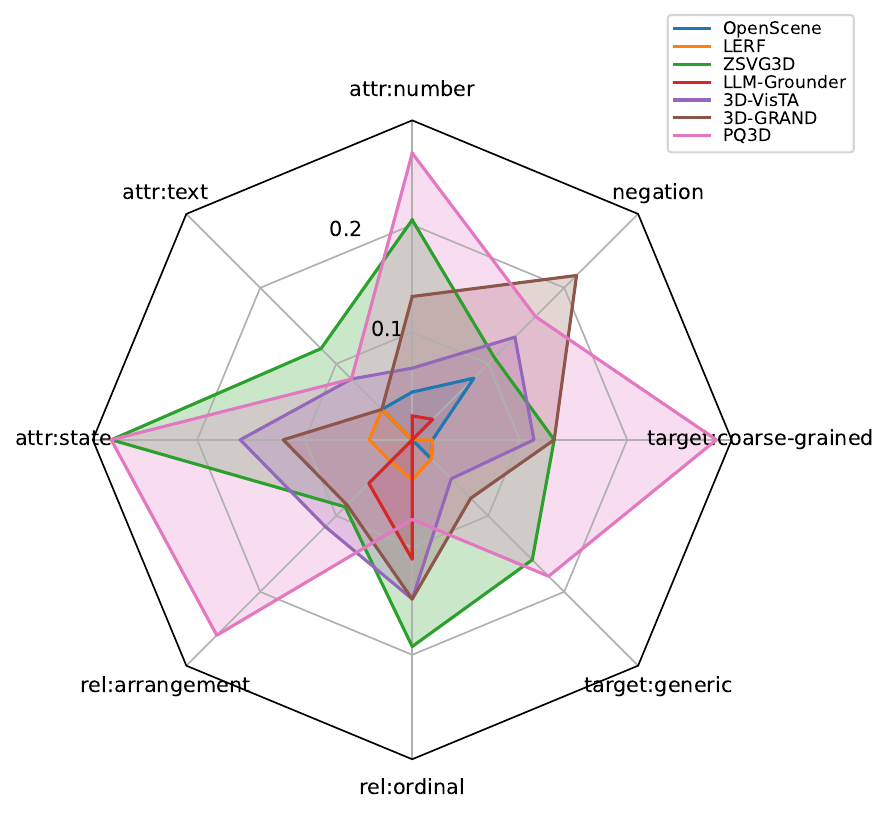}
\captionof{figure}{
  \textbf{Model Comparison on \oursshort.} We compare the performances of models on various subgroups on the ScanNet scenes of \oursshort using ground truth bounding boxes.
}
  \label{fig:model-radar}
\end{figure}

%% file: table/result_subgroup_mask3d.tex
\begin{table*}[tb]
\centering
\resizebox{\linewidth}{!}
{
\begin{tabular}{@{}l rrrrrrrrrrrrrrrrr @{}}
\toprule
& & \multicolumn{3}{c}{Attributes} & \multicolumn{4}{c}{Relationships} & \multicolumn{4}{c}{Target Reference} & \multicolumn{4}{c}{Anchor Type} & \multicolumn{1}{c}{Lang} \\
\cmidrule(){3-5} \cmidrule(l){6-9} \cmidrule(l){10-13} \cmidrule(l){14-17} \cmidrule(l){18-18}
& Overall & Num & Lab & State & Far & Arr & Ord & Comp & Gen & Cat & FG & NFN & Sing & Mul & Non & Agt & Neg \\
\midrule
OpenScene & 1.7 & 2.2 & 0.0 & 4.0 & 0.0 & 0.0 & 0.0 & 0.0 & 2.5 & 1.9 & 1.4 & 0.0 & 3.1 & 2.2 & 1.6 & 0.0 & 8.1 \\
LERF & 2.1 & 4.4 & 0.0 & 0.0 & 3.3 & 2.9 & 0.0 & 0.0 & 2.5 & 0.0 & 2.7 & 1.4 & 2.3 & 1.1 & 1.6 & 3.7 & 0.0 \\
ZSVG3D & 8.5 & 12.8 & 4.0 & \best{12.5} & 3.3 & 11.4 & 3.7 & 9.1 & 2.6 & 7.8 & 10.4 & 9.0 & 11.5 & 7.1 & 5.1 & 11.5 & 11.8 \\
LLM-Grounder & 7.1 & 8.9 & 4.0 & 8.0 & 6.7 & 8.6 & 3.7 & 4.1 & 7.5 & 3.8 & 8.2 & 11.6 & 3.8 & 6.6 & 11.5 & 3.7 & 5.4 \\
3D-VisTA & \best{15.8} & 13.3 & 4.0 & 12.0 & 13.3 & 8.6 & 3.7 & 16.3 & 7.5 & \best{9.4} & 20.4 & 14.5 & 16.0 & \best{15.4} & \best{19.7} & \best{14.8} & 8.1 \\
3D-GRAND & \best{15.8} & \best{26.7} & \best{8.0} & 12.0 & \best{16.7} & \best{14.3} & \best{11.1} & \best{22.4} & 5.0 & \best{9.4} & \best{21.1} & \best{17.4} & \best{16.8} & \best{15.4} & 16.4 & 11.1 & \best{18.9} \\
PQ3D & 10.8 & 2.2 & 4.0 & 8.0 & 3.3 & 8.6 & 3.7 & 4.1 & \best{10.0} & \best{9.4} & 11.6 & 7.2 & 10.7 & 5.5 & 8.2 & 11.1 & 8.1 \\
\bottomrule
\end{tabular}
}
\caption{\textbf{Subgroup Analysis using Mask3D}. Breakdown of accuracy at 0.25 IoU using Mask3D box predictions on \oursshort for ScanNet scenes across several subgroups of prompts.
}
\label{tab:result-subgroup-mask3d}
\end{table*}


%% file: figs/examples_supp_scannet.tex
  \begin{figure*}[t]
  \centering
  \includegraphics[trim={0 4px 0 0},clip,width=\linewidth]{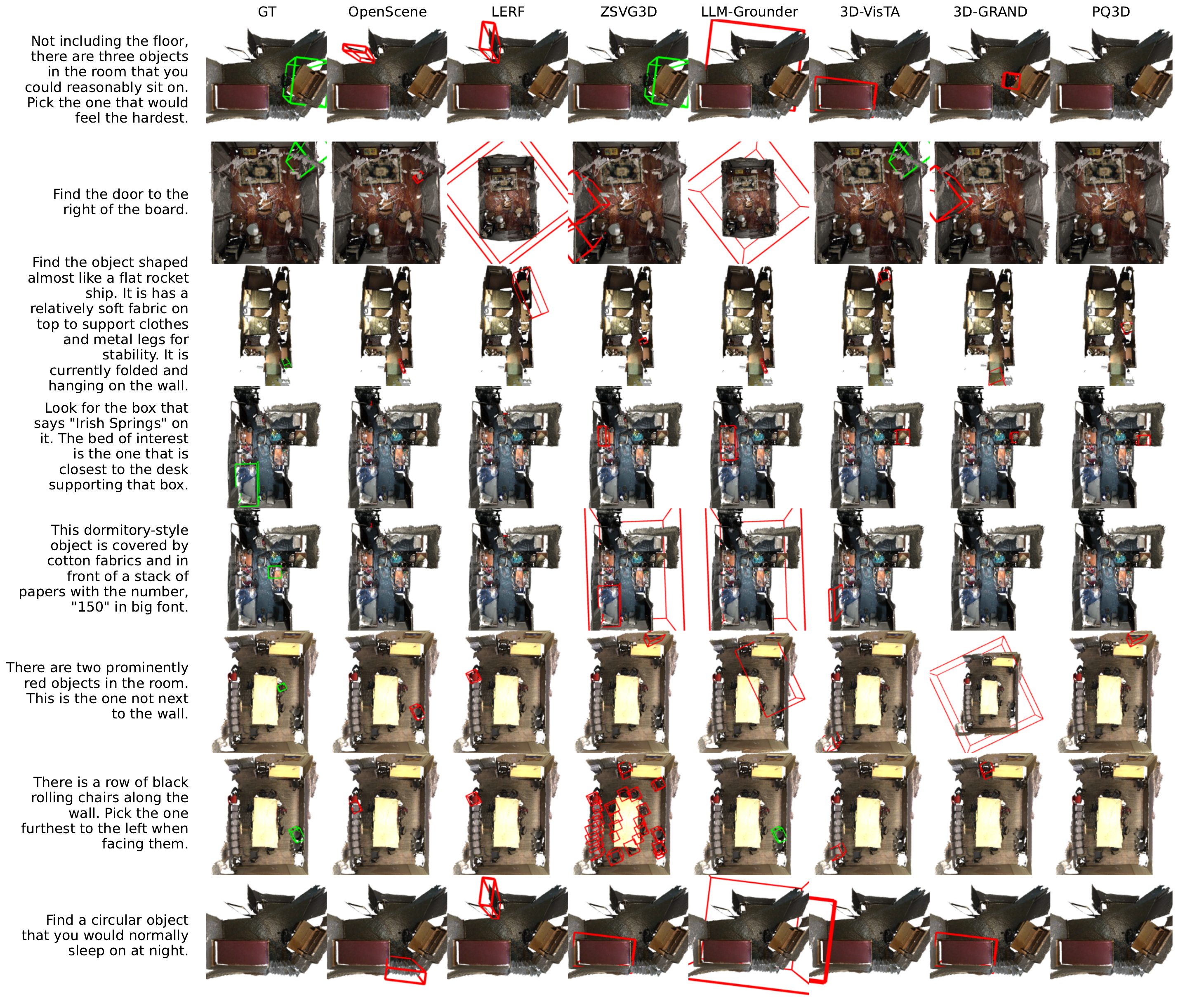}
  \vspace{-4mm}
  \caption{\textbf{Examples.} We provide additional examples for prompts from \oursshort on ScanNet scenes.
  }
  \label{fig:examples-scannet}
\end{figure*}

%% file: figs/examples_supp_scannetpp.tex
  \begin{figure*}[t]
  \centering
  \includegraphics[trim={0 4px 0 0},clip,width=\linewidth]{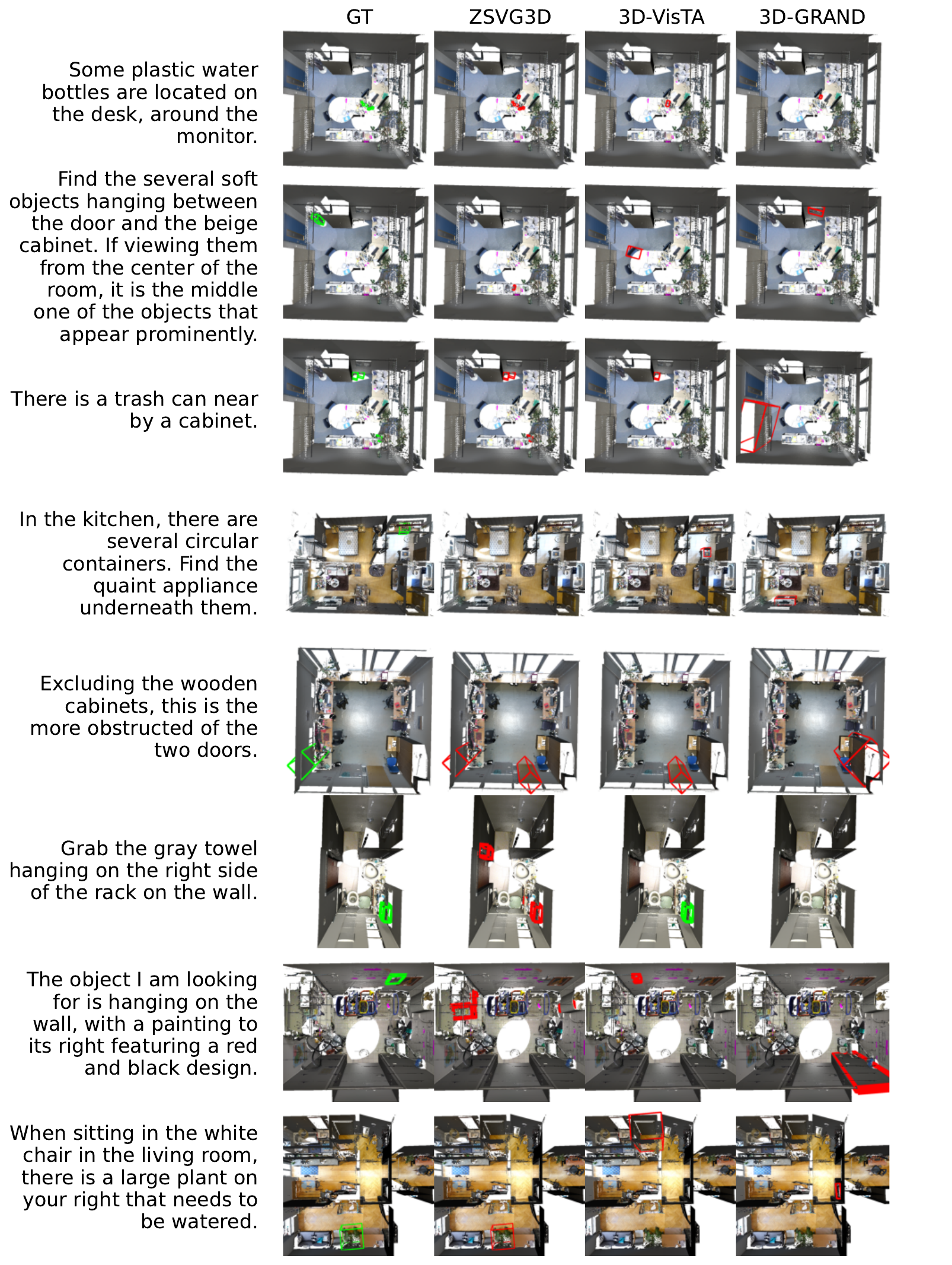}
  \vspace{-4mm}
  \caption{\textbf{Examples.} We provide additional examples for prompts from \oursshort on ScanNet++ scenes.
  }
  \label{fig:examples-scannetpp}
\end{figure*}